\documentclass[runningheads]{llncs}

\usepackage{eccv}

\usepackage{eccvabbrv}

\usepackage{graphicx}
\usepackage{booktabs}

\usepackage[accsupp]{axessibility}  % Improves PDF readability for those with disabilities.

\usepackage{array}
\usepackage{multirow}
\usepackage{caption}
\usepackage{wrapfig}
\usepackage{hyperref}

\usepackage{orcidlink}

\begin{document}

% \title{Bridging Crowd Segmentation and Counting: Semi-Supervised Exclusivity-Guided Mask Learning with Structural Consistency} 
\title{Exclusivity-Guided Mask Learning for Semi-Supervised Crowd Instance Segmentation and Counting}

\titlerunning{Exclusivity-Guided Mask Learning}

\author{Jiyang Huang\inst{1}\orcidlink{0009-0000-9291-2995} \and
Hongru Chen\inst{1} \and
Wei Lin\inst{2}\orcidlink{0000-0001-8425-956X} \and \\
Jia Wan\inst{1}\thanks{Corresponding author}\orcidlink{0000-0001-8198-1629}\and
Antoni B. Chan\inst{2}\orcidlink{0000-0002-2886-2513}}

\authorrunning{Huang et al.}

\institute{Harbin Institute of Technology, Shenzhen, China \and
City University of Hong Kong, Kowloon Tong, Hong Kong\\
\email{\{jiyanghuang0127,elonlin24,jiawan1998\}@gmail.com}\\
\email{\{24S151003\}@hit.edu.cn}\\
\email{\{abchan\}@cityu.edu.hk}}

\maketitle

\begin{abstract}
  % Crowd analysis remains a fundamental challenge in computer vision. 
  Semi-supervised crowd analysis is a prominent area of research, as unlabeled data are typically abundant and inexpensive to obtain. 
  However, traditional point-based annotations constrain performance because individual regions are inherently ambiguous, and consequently, learning fine-grained structural semantics from sparse annotations remains an unresolved challenge.
  % While crowd counting provides density estimates, crowd segmentation offers a deeper understanding by delineating individual forms; however, it poses significant challenges in dense scenarios where zero-shot foundation models often underperform. 
  In this paper, we first propose an Exclusion-Constrained Dual-Prompt SAM (EDP-SAM), based on our Nearest Neighbor Exclusion Circle (NNEC) constraint, to generate mask supervision for current datasets.
  % In this paper, we propose a novel Semi-Supervised Region-Aware Discriminative Crowd Segmentation (SS-RDCS) method, based on our NNEC (nearest neighbor exclusion circle) restriction Dual-Prompt SAM (NDP-SAM) auto-generated mask dataset. 
  With the aim of segmenting individuals in dense scenes, we then propose Exclusivity-Guided Mask Learning (XMask), which enforces spatial separation through a discriminative mask objective. Gaussian smoothing and a differentiable center sampling strategy are utilized to improve feature continuity and training stability. 
  % Specifically, we introduce an Adaptive region Partitioning approach combined with Gaussian smoothing embedding and a differentiable center sampling strategy to enhance feature continuity and training stability. 
  Building on XMask, we present a semi-supervised crowd counting framework that uses instance mask priors as pseudo-labels, which contain richer shape information than traditional point cues. 
  % Significantly, we create a semi-supervised Mask Constraint Counting Optimization (MCCO) strategy, which leverage the pseudo-masks generated generated via SS-RDCS with selection mechanism and a set of mask-constrained loss formulation, not only validating the segmentation quality but also yielding substantial improvements in counting precision. 
  Extensive experiments on the ShanghaiTech A, UCF-QNRF, and JHU++ datasets (using 5\%, 10\%, and 40\% labeled data) verify that our end-to-end model achieves state-of-the-art semi-supervised segmentation and counting performance, effectively bridging the gap between counting and instance segmentation within a unified framework. Code can be found at https://github.com/JoyceeH0127/ECCV2026XMask.
  
  \keywords{Semi-supervised learning \and Crowd Counting \and Instance Segmentation}
\end{abstract}

\section{Introduction}
\label{sec:intro}

Crowd scene analysis has long been a pivotal research topic in computer vision, driven by its critical applications in public safety, smart city management, and traffic monitoring. Despite significant progress in density estimation and localization \cite{50-Zhang2016SingleImageCC,51-idrees2018compositionlosscountingdensity,56-li2018csrnetdilatedconvolutionalneural,13-ma2019bayesianlosscrowdcount,24-song2021rethinkingcountinglocalizationcrowdsa}, the field remains constrained by a severe data annotation bottleneck. The prohibitive cost of acquiring pixel-level ground truth forces a reliance on point-based annotations, which inherently lack structural semantics regarding individual shape and occupancy. 

Current crowd analysis research spans multiple directions: density regression and point-based localization, which treat individuals as dimensionless points, ignoring spatial extent; and object detection research \cite{57-stewart2015endtoendpeopledetectioncrowded,58-liu2019adaptivenmsrefiningpedestrian,59-wang2018repulsionlossdetectingpedestrians,60-chu2020detectioncrowdedscenesproposal,61-ge2020psrcnndetectingsecondaryhuman} struggles with bounding box regression in scenarios of severe occlusion. Other related methods \cite{62-liu2019pointinboxout,63-sam2020locatesizecountaccurately} connect the detection task with counting, but lack in semantic information. While recent foundation models like SAM \cite{5-kirillov2023segment} demonstrate zero-shot segmentation potential, their direct application is hindered by high computational latency and performance degradation in dense, ambiguous regions \cite{39-cai2024crowdsamsamsmartannotator}. Consequently, learning fine-grained structural semantics from sparse annotations remains an open problem. In particular, how to harness additional data to compensate for the lack of dense supervision has not been sufficiently investigated.

\begin{figure}
    \centering
    \vspace{-5mm}
    \includegraphics[width=1.0\linewidth]{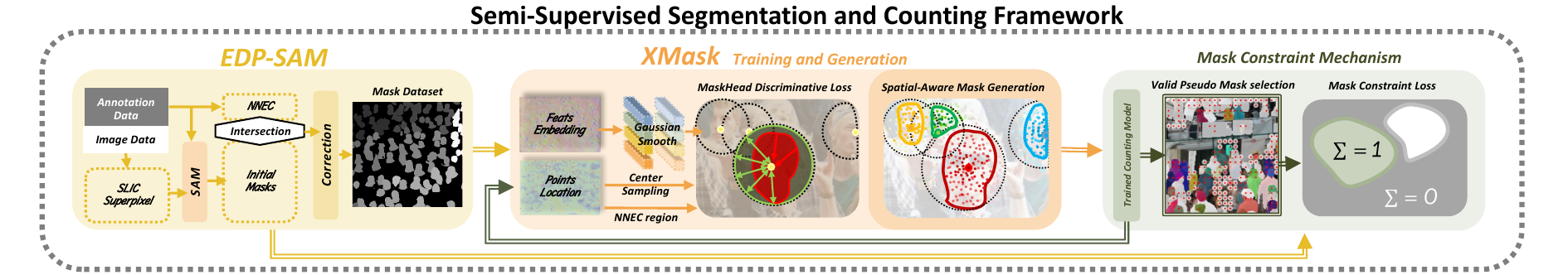}
    \caption{Conception of our approach. EDP-SAM is designed to generate data with mask supervision. To acquire segmentation capabilities, we propose a semi-supervised XMask method to learn a discriminative feature representation. Subsequently, we present a mask-constraint strategy composing the final step in our framework, coupling semantic and spatial information to enhance counting accuracy. }
    \label{fig:contribution}
    \vspace{-5mm}
\end{figure}

To address these limitations, we propose a novel paradigm that extends crowd counting to dense instance segmentation without relying on expensive point and mask annotations, as illustrated in \cref{fig:contribution}. Our objective is to achieve high instance segmentation performance while enhancing counting accuracy in dense crowds. First, we employ SAM\cite{5-kirillov2023segment} to generate datasets with masks by using existing annotations as point prompts and SLIC \cite{73-10.1109/TPAMI.2012.120}-generated superpixels for region guidance. The \textbf{N}earest-\textbf{N}eighbor \textbf{E}xclusion \textbf{C}ircle \textbf{(NNEC)} is further introduced as a spatial constraint to suppress interference from adjacent instances. This design, \textbf{E}xclusion-Constrained \textbf{D}ual-\textbf{P}rompt \textbf{S}AM \textbf{(EDP-SAM)}, yields spatially consistent and reliable pseudo-masks. After manual correction, the resulting datasets are subsequently used for model training and standardized comparison with baseline methods.

% Accrodingly, we introduce a \textbf{S}emi-\textbf{S}upervised \textbf{R}egion-Aware \textbf{D}iscriminative \textbf{C}rowd \textbf{S}egmentation \textbf{(SS-RDCS)} methods.
We then propose an E\textbf{X}clusivity-Guided \textbf{Mask} Learning \textbf{(XMask)} for crowd instance segmentation by enforcing spatial separation through a discriminative mask objective. First, we propose NNEC, a metric-induced local topological partition that guarantees first-order exclusivity under sparse supervision. Next, by leveraging discriminative feature embedding learning \cite{33-8803021,64-debrabandere2017semanticinstancesegmentationdiscriminative,66-Chen_2024}, we reformulate the instance segmentation task as a semantic segmentation problem within dynamically defined local regions, which significantly enhances computational efficiency. Additionally, to improve feature continuity and training stability, we incorporate Depthwise Gaussian smoothing into the embedding space and use a Differentiable Center Sampling strategy to ensure gradient consistency.

Based on the proposed crowd instance segmentation architecture, we further design a semi-supervised mask-constraint mechanism that provides richer guidance than point labels for semi-supervised crowd counting. By incorporating a set of mask constraint losses and coupling them with a valid mask filtering strategy, the proposed mechanism forms a feedback loop that enhances the reliability and consistency of pseudo-mask supervision. 
% Related approach Topological constraint in \cite{42-Abousamra2020LocalizationIT} is similar to our method, but computes similarity between dot-square masks, while ours utilizes semantic masks and directly processes the predicted point map 
Rather than leveraging feature embeddings to generate instance pseudo-masks for enforcing a strict spatial containment, Abousamra, Shahira et al. \cite{42-Abousamra2020LocalizationIT} impose a topological constraint on the likelihood map's landscape using persistence loss, ensuring the number of connected components in the predicted heatmap corresponds exactly to the number of ground truth dots. Our design allows the segmentation branch to regularize crowd counting in a semi-supervised manner. Crucially, our specialized framework delivers segmentation performance on par with foundation models (e.g., SAM) but with drastically reduced computational costs, presenting a robust and efficient alternative for dense crowd analysis. Extensive experiments on the ShanghaiTech A (ShTech A), UCF-QNRF, and JHU++ datasets (using 5\%, 10\%, and 40\% labeled protocols) demonstrate that our end-to-end model significantly outperforms point-prompt FastSAM\cite{49-zhao2023fastsegment}-based methods in both inference speed and segmentation accuracy.

The contributions of our paper are summarized as follows:

% \begin{itemize}
%     \item We propose a novel prompt-based mask generation pipeline that leverages location points and SLIC superpixels to guide the \textit{SAM}. By incorporating NNEC regional constraints, this strategy effectively segments individuals in dense crowds. Furthermore, we construct a large-scale crowd mask dataset to facilitate future research.

%     \item We present SS-RDCS, a specialized instance segmentation framework tailored for dense crowd scenes. Key technical innovations include a MaskHead Discriminative Loss, a Gaussian Smoothing strategy with grid sampling to refine feature embedding learning and spatial and regional constraint mechanisms to regularize mask generation and training.

%     \item Building upon our segmentation model, we design a robust three-stage semi-supervised training framework for crowd counting, introducing a mask constraint loss and a high-confidence mask selection mechanism and enhancing the counting accuracy.

%     \item Our method achieves state-of-the-art results, effectively bridging the gap between crowd counting and segmentation. Compared to existing FastSAM-based approaches, our framework demonstrates superior efficiency and higher recall in segmentation while maintaining precise counting performance.
    
% \end{itemize}

\begin{itemize}
    \item We propose Exclusion-Constrained Dual-Prompt SAM (EDP-SAM), an automatic mask generation framework for dense crowds that guides SAM with head-point and superpixel dual prompts while enforcing a Nearest Neighbor Exclusion constraint to prevent cross-instance leakage. This design improves individual separation in highly congested scenes, and the resulting pseudo-masks are lightly refined to build a high-quality crowd segmentation dataset.
    % \item NNEC restricted Dual Prompt SAM (NDP-SAM): an auto-mask generation strategy that leverages head point and superpixel dual-prompt to guide the SAM \cite{5-kirillov2023segment}, incorporating Nearest Neighbor Exclusion Circle (NNEC) regional constraints. This strategy effectively segments individuals in dense crowds, and we construct crowd datasets with mask annotations after manual correction.

    \item We introduce Exclusivity-Guided Mask Learning (XMask), an exclusivity-guided semi-supervised method for dense crowd instance segmentation that enforces spatial separation through a discriminative mask objective, coupled with Gaussian smoothing and differentiable center sampling for robust learning under limited supervision.
    % \item SS-RDCS: a semi-supervised instance segmentation framework tailored for dense crowd scenes. Key technical innovations comprise a MaskHead Discriminative Loss integrated with regional-spatial constraint, alongside a robust regularization framework employing Gaussian smoothing and grid-based sampling.

    \item Building on XMask, we further introduce an instance mask prior framework for semi-supervised crowd counting, in which reliable instance masks replace point annotations as structured pseudo-labels. By leveraging richer spatial and shape information instead of sparse point cues, the framework produces more accurate pseudo-labels and significantly enhances the performance.
    % \item Building upon XMask, we then propose instance mask priors for semi-supervised crowd counting with mask supervision and a high-confidence mask selection mechanism . 

    \item Our semi-supervised strategy achieves state-of-the-art results, effectively bridging the gap between crowd counting and segmentation, constructing feature structural consistency, demonstrating superior segmentation efficiency and higher recall while maintaining precise counting performance.
    
\end{itemize}

\section{Related Works}

\subsection{Instance Segmentation}
Instance segmentation is a core foundational task in computer vision. It not only requires detecting all objects of interest in an image but also assigns precise category labels and instance IDs to each object's pixels.

\textbf{Crowd Instance Segmentation.} 
Crowd instance segmentation aims to separate each individual in highly congested scenes, where severe occlusion and overlap make precise delineation challenging. Early deep learning approaches \cite{41-kang2014fullyconvolutionalneuralnetworks} relied on fully convolutional networks to directly predict crowd regions. With the rise of foundation models such as SAM \cite{5-kirillov2023segment} and FastSAM \cite{49-zhao2023fastsegment}, recent efforts have shifted toward adapting prompt-driven segmentation frameworks to dense small-object scenarios. For instance, CrowdSAM \cite{39-cai2024crowdsamsamsmartannotator} utilizes density-derived prompts and post-processing refinement to enhance mask quality. Nevertheless, limitations in inference efficiency and scalability remain unresolved.

% The crowd instance segmentation task requires identifying and segmenting each individual within crowd, which is particularly challenging in extremely crowded scenes due to high overlap between individuals. One of the earliest deep learning approaches to crowd segmentation \cite{41-kang2014fullyconvolutionalneuralnetworks} employed a fully Convolutional architecture to directly predict crowd region segmentation maps, optimized using dynamic image information and motion features. With the emergence of large visual models like SAM\cite{5-kirillov2023segment} and FastSAM\cite{49-zhao2023fastsegment}, research in crowd segmentation has focused on leveraging SAM\cite{5-kirillov2023segment} powerful zero-shot capabilities and adapting it to dense small-object scenarios through fine-tuning or adapters. CrowdSAM\cite{39-cai2024crowdsamsamsmartannotator} intelligently filters key points from predicted density maps as prompts and employs post-processing discrimination to achieve effective crowd segmentation. However, issues such as limited inference speed persist.

\textbf{Two Stage Instance Segmentation.} Two-stage instance segmentation follows a proposal-to-refinement pipeline: it first generates candidate regions and then refines them into final masks. In this framework, methods are mainly categorized into top-down and bottom-up approaches \cite{32-10.1016/j.neucom.2025.129584}.

\textbf{Top-down} instance segmentation methods rely on high-performance detectors to generate RoIs, followed by independent mask prediction within each region, effectively decomposing the task into local binary segmentation. Mask R-CNN \cite{26-he2018maskrcnn} established this paradigm by adding a parallel mask branch to Faster R-CNN \cite{25-ren2016fasterrcnnrealtimeobject}. Subsequent works improved feature propagation and mask quality, including PANet \cite{27-Liu_2018_CVPR} with bottom-up path aggregation, Cascade Mask R-CNN \cite{28-ding2021deeplyshapeguidedcascadeinstance} with multi-stage refinement, and several boundary-aware extensions \cite{29-cheng2020boundarypreservingmaskrcnn,30-zhang2021refinemaskhighqualityinstancesegmentation,31-ke2021masktransfinerhighqualityinstance} that enhance pixel-level mask accuracy.

% The core idea of \textbf{top-down} methods is to leverage existing high-performance object detectors to generate high-quality region of interest candidates, then independently perform mask prediction within each RoI. This simplifies the instance segmentation problem into a series of local binary classification tasks. Mask R-CNN \cite{26-he2018maskrcnn} stands as a landmark achievement in this field. It adds a parallel mask prediction branch to Faster R-CNN \cite{25-ren2016fasterrcnnrealtimeobject}, operating alongside the classification and regression branches. PANet \cite{27-Liu_2018_CVPR} addressed the issue of excessively long feature propagation paths in FPN lower layers, it introduces bottom-up path aggregation. Cascade Mask R-CNN\cite{28-ding2021deeplyshapeguidedcascadeinstance} directly extends the concept by cascading multiple detection heads and mask heads, progressively increasing the IoU threshold. Furthermore, regarding the coarse edge issue in Mask R-CNN \cite{26-he2018maskrcnn}, a series of works also aim to enhance the pixel-level quality of masks \cite{29-cheng2020boundarypreservingmaskrcnn,30-zhang2021refinemaskhighqualityinstancesegmentation,31-ke2021masktransfinerhighqualityinstance}.

\textbf{Bottom-up} approaches \cite{34-liang2021instancesegmentation3dscenes,35-8237640} learn pixel-wise embeddings instead of relying on predefined anchors, encouraging pixels of the same instance to cluster in a high-dimensional space while separating different instances. This paradigm is typically optimized via discriminative losses with pull–push forces. Representative works \cite{34-liang2021instancesegmentation3dscenes,33-8803021,64-debrabandere2017semanticinstancesegmentationdiscriminative,66-Chen_2024}jointly predict semantic maps and pixel embeddings to simplify training, while others \cite{36-8099788,37-ijcai2021p175} exploit spatial affinity cues for grouping. Embedding-based methods have shown particular advantages in irregular scenarios, such as biomedical images, as demonstrated by EmbedSeg \cite{38-LALIT2022102523}.

% The \textbf{bottom-up} approach  \cite{34-liang2021instancesegmentation3dscenes,35-8237640} no longer relies on predefined anchor boxes but instead learns to map each pixel in an image to a high-dimensional embedding space. Within this space, pixels belonging to the same instance cluster closely together, while pixels from different instances remain distant. Discriminative loss and affinity learning form the theoretical foundation of this method. Discriminative loss, mentioned in \cite{33-8803021,64-debrabandere2017semanticinstancesegmentationdiscriminative,66-Chen_2024}, defines a loss function incorporating pull and push forces. The approach by Newell et al. \cite{34-liang2021instancesegmentation3dscenes} simultaneously outputs semantic segmentation results and embedding labels for pixels, significantly simplifying the training objective. Beyond directly learning embedding distances, another approach leverages the spatial distribution characteristics of pixels \cite{36-8099788,37-ijcai2021p175}. For irregularly shaped biomedical images, EmbedSeg \cite{38-LALIT2022102523} demonstrated the superiority of embedding-based methods.

\textbf{Semi-supervised Instance Segmentation.} This task learns from scarce annotations and abundant unlabeled data, commonly relying on pseudo-labeling or consistency constraints to exploit unlabeled samples.

Early semi-supervised instance segmentation methods were primarily built upon anchor-based frameworks such as Mask R-CNN \cite{26-he2018maskrcnn}. Noisy Boundaries \cite{71-Wang2022NoisyBL} and PAIS \cite{2-hu2023pseudolabelalignmentsemisupervisedinstance} enhanced pseudo-label robustness via boundary-aware modeling and dynamic alignment, respectively, yet remained limited by architectural complexity or label reweighting strategies. To adapt to anchor-free paradigms, Polite Teacher \cite{3-filipiak2022politeteachersemisupervisedinstance} adopted a CenterMask-style design with mutual learning, improving efficiency but showing weakness on extremely small objects. More recently, Guided Distillation \cite{4-errada2023guideddistillationsemisupervisedinstance} introduced Vision Transformers to leverage pretrained representations, albeit with increased memory overhead.

% Early methods were predominantly based on anchor-based detectors.  Noisy Boundaries\cite{71-Wang2022NoisyBL} employs a Mask R-CNN \cite{26-he2018maskrcnn} architecture, proposing Noise-Tolerant Masking (NTM) + Boundary Protection Mapping. This approach demonstrates strong robustness against boundary noise but incurs higher model complexity; PAIS\cite{2-hu2023pseudolabelalignmentsemisupervisedinstance} also utilizes the same underlying architecture but incorporates a dynamic alignment loss to address the issue of inconsistent pseudo-label quality, but it only optimizes the label weights; Polite Teacher\cite{3-filipiak2022politeteachersemisupervisedinstance} adopts a Center Mask architecture and employs a mutual learning and mask score filtering approach to better adapt to anchor-free architectures, achieving accelerated inference, while facing limitations in detecting extremely small objects. Guided Distillation\cite{4-errada2023guideddistillationsemisupervisedinstance} systematically introduces Vision Transformers into a semi-supervised instance segmentation framework for the first time, fully leveraging pre-trained models. However, due to the presence of ViT architecture, it imposes high demands on graphics memory.

The emergence of vision foundation models like SAM \cite{5-kirillov2023segment} has shifted semi-supervised instance segmentation toward prompt-driven and more generalizable paradigms, improving pseudo-label quality. Methods like S\^{}4M\cite{6-yoon2025s4mboostingsemisupervisedinstance} and SemiSAM+\cite{7-zhang2025semisamrethinkingsemisupervisedmedical} exploit SAM \cite{5-kirillov2023segment} as a powerful teacher to refine mask supervision. Meanwhile, diffusion-based approaches\cite{8-ulmer2025conditionallatentdiffusionmodels,1-wang2022noisyboundarieslemonlemonade,10-diffusion}, including DiffusionInst \cite{9-gu2022diffusioninstdiffusionmodelinstance}, model segmentation as a denoising process and leverage generative modeling for semantic alignment. Despite enhanced robustness to ambiguous boundaries, these methods often suffer from substantial computational overhead in dense scenarios.

% The emergence of vision foundation models like SAM \cite{5-kirillov2023segment} has redefined the landscape of instance segmentation, steering the field beyond traditional semi-supervised formulations toward promptable and generalizable segmentation paradigms, offering a new approach to addressing the issue of pseudo-label quality. S\^{}4M\cite{6-yoon2025s4mboostingsemisupervisedinstance} and SemiSAM+\cite{7-zhang2025semisamrethinkingsemisupervisedmedical} are typical examples, employing SAM \cite{5-kirillov2023segment} as a “super teacher” compensates for the shortcomings of traditional teaching models in peripheral details. Furthermore, semi-supervised methods in diffusion models demonstrate greater robustness than traditional discriminative models when handling ambiguous boundaries and complex textures\cite{8-ulmer2025conditionallatentdiffusionmodels}. DiffusionInst\cite{9-gu2022diffusioninstdiffusionmodelinstance} formulates instance segmentation as a denoising process from noise to filter. Diffusion-based SSL\cite{1-wang2022noisyboundarieslemonlemonade,10-diffusion} leverages the generative capacity of diffusion models to align the semantic distributions in latent spaces. Nevertheless, their practical deployment is often constrained by substantial computational overhead, especially in dense and highly overlapping scenarios.

\subsection{Semi-supervised Crowd Counting}
Semi-Supervised Crowd Counting is a specialized sub-field of computer vision that addresses the challenge of estimating the number of individuals in dense crowds using a limited amount of labeled data alongside a large corpus of unlabeled data.

\textbf{Density-Based Counting.} Crowd counting annotation is costly and labor-intensive. Recent Works\cite{11-ssc-multi-task,12-liu2020semisupervisedcrowdcountingselftraining,13-ma2019bayesianlosscrowdcount} explore semi-supervised learning to leverage unlabeled data for improved counting performance. Liu et al. \cite{14-liu2018leveragingunlabeleddatacrowd,15-Liu_2019} propose a Learning-to-Rank (L2R) module that models the ordinal relationship between image patches and their sub-patches. DACount \cite{16-lin2022semisupervisedcrowdcountingdensity} and P³Net \cite{17-lin2024semisupervisedcountingpixelbypixeldensity} improve density discrimination by assigning density-level labels to image patches. Approaches \cite{18-meng2021spatialuncertaintyawaresemisupervisedcrowd,19-10.1109/TCSVT.2023.3241175} incorporate pixel-wise uncertainty to suppress noise in pseudo density maps, promoting pseudo-label quality.

\textbf{Localization-Based Counting.} Semi-supervised localization-based counting methods generally adopt a teacher–student framework. OT-M \cite{22-Lin_2023_CVPR} derives pseudo-points from predicted density maps using optimal transport. CU \cite{23-li2023calibratinguncertaintysemisupervisedcrowd}, built upon P2PNet \cite{24-song2021rethinkingcountinglocalizationcrowdsa}, improves pseudo-label reliability by selecting high-confidence patches based on uncertainty estimation. SAL \cite{70-10.1145/3664647.3680976} incorporates active learning to dynamically divide data, treating difficult samples as labeled and easier ones as unlabeled. P2R \cite{21-lin2025pointtoregionlosssemisupervisedpointbased} introduces a points to region supervision strategy that segments regions around pseudo-points to propagate confidence more effectively than traditional P2PNet\cite{24-song2021rethinkingcountinglocalizationcrowdsa} matching. Consistent-Point\cite{20-zou2025consistent} improves semi-supervised point-prompt crowd counting by refining pseudo-points through position aggregation and uncertainty calibration for more consistent supervision.

\section{Methods}
\begin{figure}[!t]
    \centering
    \includegraphics[width=1.0\linewidth]{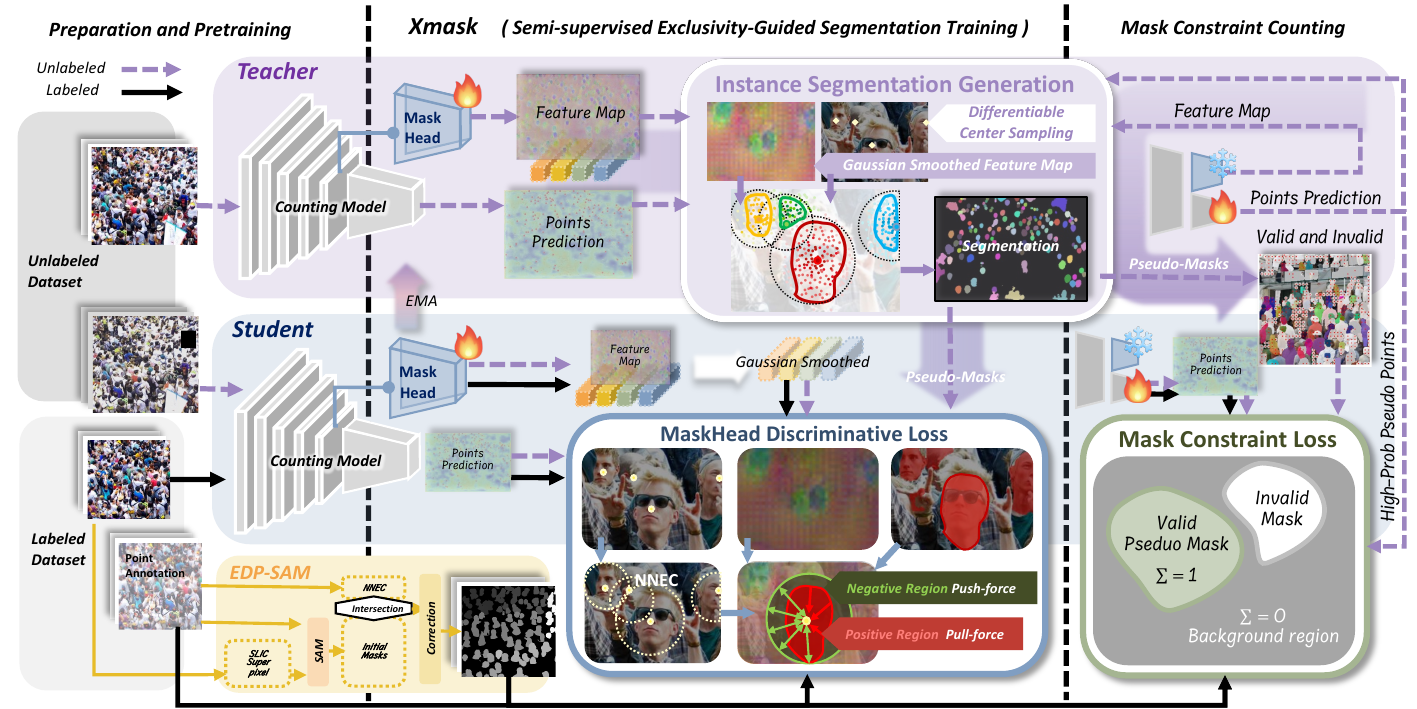}
    \caption{Overview of the proposed framework. Our semi-supervised pipeline consists of three stages: (1) pretraining and mask preparation, (2) crowd segmentation training, and (3) mask-constrained crowd counting optimization. In Stage 1, mask annotations are generated from point labels via the proposed EDP-SAM (yellow region). Stage 2 presents the semi-supervised XMask training process, including discriminative supervision and pseudo-mask generation. The trained segmentation model is then utilized in Stage 3 to produce pseudo-masks for unlabeled data, where newly designed loss functions are introduced to further enhance counting performance via mask constraint.}
    \label{fig:3-stage-framework}
    \vspace{-5mm}
\end{figure}

To ensure robust semi-supervised instance segmentation performance in dense crowds, we a propose framework that consists of three stages.
% This section details our method for semi-supervised crowd instance segmentation and counting. 
As shown in \cref{fig:3-stage-framework}, our model establishes a synergistic bridge between semi-supervised instance segmentation and counting within a single end-to-end architecture, enabling the two tasks to mutually reinforce each other in a closed-loop manner. 
The proposed method achieves state-of-the-art semi-supervised crowd instance segmentation and counting performance.

Our approach builds on P2R\cite{21-lin2025pointtoregionlosssemisupervisedpointbased}, which mitigates noisy pseudo-label locations by matching predictions within a neighborhood. Structurally, we streamline the P2PNet \cite{24-song2021rethinkingcountinglocalizationcrowdsa} architecture by removing the regression branch in favor of pixel grid coordinates. Our model incorporates a feature backbone, fusion neck, and classification head, alongside an innovative introduced parallel mask head for segmentation. For semi-supervised learning, we adopt the Mean Teacher paradigm\cite{44-tarvainen2018meanteachersbetterrole}, where the student optimizes joint losses while the teacher generates pseudo-labels, which follows standard procedure \cite{47-Lee2013PseudoLabelT,48-sohn2020fixmatchsimplifyingsemisupervisedlearning}, with parameters updated via the student exponential moving average (EMA) \cite{45-cai2021exponentialmovingaveragenormalization,46-moralesbrotons2024exponentialmovingaverageweights}. The whole pipeline can be trained end-to-end.

\subsection{Dataset Construction with Dense Mask Supervision }

We introduce EDP-SAM to automatically generate and manually verify a set of high-quality masks, providing essential structural guidance for our segmentation-counting dual-tasks optimization. % The specific implementation follows.

\begin{figure}[!t]
    % \centering
    \includegraphics[width=1.0\linewidth]{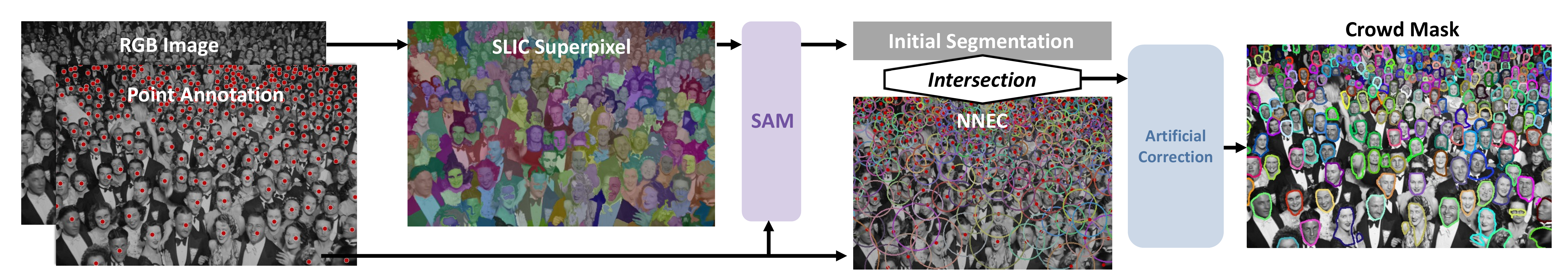}
    \caption{Point-Superpixel Dual-Prompt SAM with NNEC Constraint Method}
    \label{fig:Mask dataset}
    \vspace{-5mm}
\end{figure}

Since existing crowd counting datasets only contain point annotation data $\mathbf{p}_i = (y_i, x_i)$, our primary task is to construct the corresponding mask supervision datasets. Inspired by relevant research \cite{67-jimaging11060172,68-wang2023samrsscalingupremotesensing,69-Ma_2024}, we also employed the SAM model for automatic construction, supplemented by partial manual correction. As illustrated in \cref{fig:Mask dataset}, we fed both the original annotation points and superpixels generated by SLIC \cite{73-10.1109/TPAMI.2012.120} as dual-prompts into SAM. We then intersected the segmentation output from SAM with the Nearest-Neighbor Exclusion Circle (NNEC) regions corresponding to each point. The reason for our design will be illustrated in the supplementary materials.

Specifically, the purpose of applying NNEC constraint is to identify the maximum mask area corresponding to each annotation, as the SAM often struggles to segment individual shapes in dense scenes, even when specific point prompts are provided. The specific calculation of NNEC radius is as follows: 
\begin{equation}
    r_i =
    \displaystyle \min_{j \neq i} \lVert \mathbf{p}_i - \mathbf{p}_j \rVert_2,
    \label{nnec radius}
\end{equation}
where $\mathbf{p}_i$ and $\mathbf{p}_j$ are two point prompts indicating two head locations. Based on EDP-SAM and manual correction, we generate mask annotations for major crowd counting datasets for future crowd instance segmentation research.

\subsection{Semi-supervised Crowd Instance Segmentation}
\label{instance segmentation method}
Based on the mask annotations, we propose Exclusivity-Guided Mask Learning (XMask), a semi-supervised method for dense crowd instance segmentation that enforces spatial separation through a discriminative mask objective, coupled with Gaussian smoothing and differentiable center sampling for robust learning under limited supervision.
% This section concentrates on our XMask method, including the formulation of semi-supervised segmentation training, the preprocessing procedures, and the inference mechanism. 
As shown in \cref{fig:3-stage-framework}, after obtaining a preliminary counting model, we train an instance segmentation network within the same semi-supervised framework using the proposed XMask. 
% The original network architecture was frozen, with training confined solely to the maskhead component. 

\subsubsection{MaskHead Discriminative Loss} 
\label{MaskHead Discriminative loss.}
\begin{figure}[!t]
    \centering
    \includegraphics[width=\linewidth]{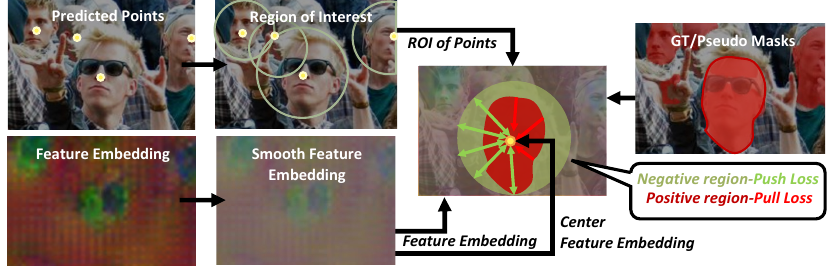}
    \caption{MaskHead Discriminative Loss. First, use points to obtain the corresponding ROIs for computation. Based on the pseudo/ground truth mask, divide the ROIs into positive and negative regions. After applying Gaussian smoothing, extract the center feature and compute pixel-level L2 distance within feature map. Finally, applying pull and push force losses separately.}
    \label{fig:discriminative loss}
\end{figure}

This method proposes a feature embedding learning loss function for point-supervised instance segmentation. Unlike traditional per-pixel classification paradigms, we introduce NNEC adaptive radius local circular region sampling and asymmetric marginal contrast constraints within the pixel embedding space. The whole process is visualized in \cref{fig:discriminative loss}.

Specifically, given an input image $\mathbf{I} \in \mathbb{R}^{H \times W \times 3}$  and its corresponding embedding feature map $\mathbf{E} \in \mathbb{R}^{D \times H_e \times W_e}$ where D is the feature dimension, a set of model predict point annotations $\mathcal{P} = \{\mathbf{p}_i\}_{i=1}^{N}$ where $\mathbf{p}_i = (y_i, x_i)$ denotes the coordinates of annotated points, and instance-level ground truth masks $\mathbf{M} \in \mathbb{Z}_{\ge 0}^{H \times W}$. 
% First, bilinear interpolation is used to up-sample the embedded feature map to match the spatial dimensions of the original size, $\mathbf{\hat{E}} \in \mathbb{R}^{D \times H \times W}$.

\textbf{Depthwise Gaussian Smoothing.} To mitigate high-frequency interpolation artifacts and local noise in sampled feature maps, we employ a separable depthwise Gaussian blur that enhances spatial coherence without compromising embedding discriminability and the effectiveness is demonstrated in \cref{sec:ablation}. Given a kernel size $k$ and standard deviation $\sigma$, we construct discrete coordinates $u$ and a normalized one-dimensional Gaussian kernel:

\begin{equation}
    g(u) =
    \frac{
    \exp\!\left(-\frac{u^2}{2\sigma^2}\right)
    }{\sum_{u'}\exp\!\left(-\frac{{u'}^2}{2\sigma^2}\right)},
    \quad
    u \in \left[-\left\lfloor \frac{k}{2} \right\rfloor, \dots, \left\lfloor \frac{k}{2} \right\rfloor \right].
  \label{eq:one-dimensional Gaussian kernel}
\end{equation}

Utilizing the separable property of two-dimensional isotropic Gaussian functions $G(u,v) = g(u)\, g(v)$,  we decompose a two-dimensional convolution into a cascade of two one-dimensional convolutions:

\begin{equation}
    \hat{E} * G = (\hat{E} * g_x) * g_y.
    \label{eq:two one-dimensional  convolutions}
\end{equation}

Next, depthwise convolution is applied, enabling the same Gaussian kernel to be applied independently to each channel. Specifically, the smoothed embedding feature map $\tilde{E}$ is obtained by performing channel-wise convolutions sequentially along the horizontal and vertical directions \cref{eq:horizontal conv,eq:vertical conv}, and the convolutional weights $W$ are fixed to $g$.

\begin{equation}
    \tilde{E}^{(1)} =
    \mathrm{DWConv2d}
    \big(
    \tilde{E},
    \; W_x \in \mathbb{R}^{D \times 1 \times 1 \times k},
    \; \text{pad}=(0,\lfloor k/2 \rfloor)
    \big),
    \label{eq:horizontal conv}
\end{equation}

\begin{equation}
    \tilde{E}' =
    \mathrm{DWConv2d}
    \big(
    \tilde{E}^{(1)},
    \; W_y \in \mathbb{R}^{D \times 1 \times k \times 1},
    \; \text{pad}=(\lfloor k/2 \rfloor, 0)
    \big).
    \label{eq:vertical conv}
\end{equation}

\begin{figure}
    \centering
    \includegraphics[width=0.8\linewidth]{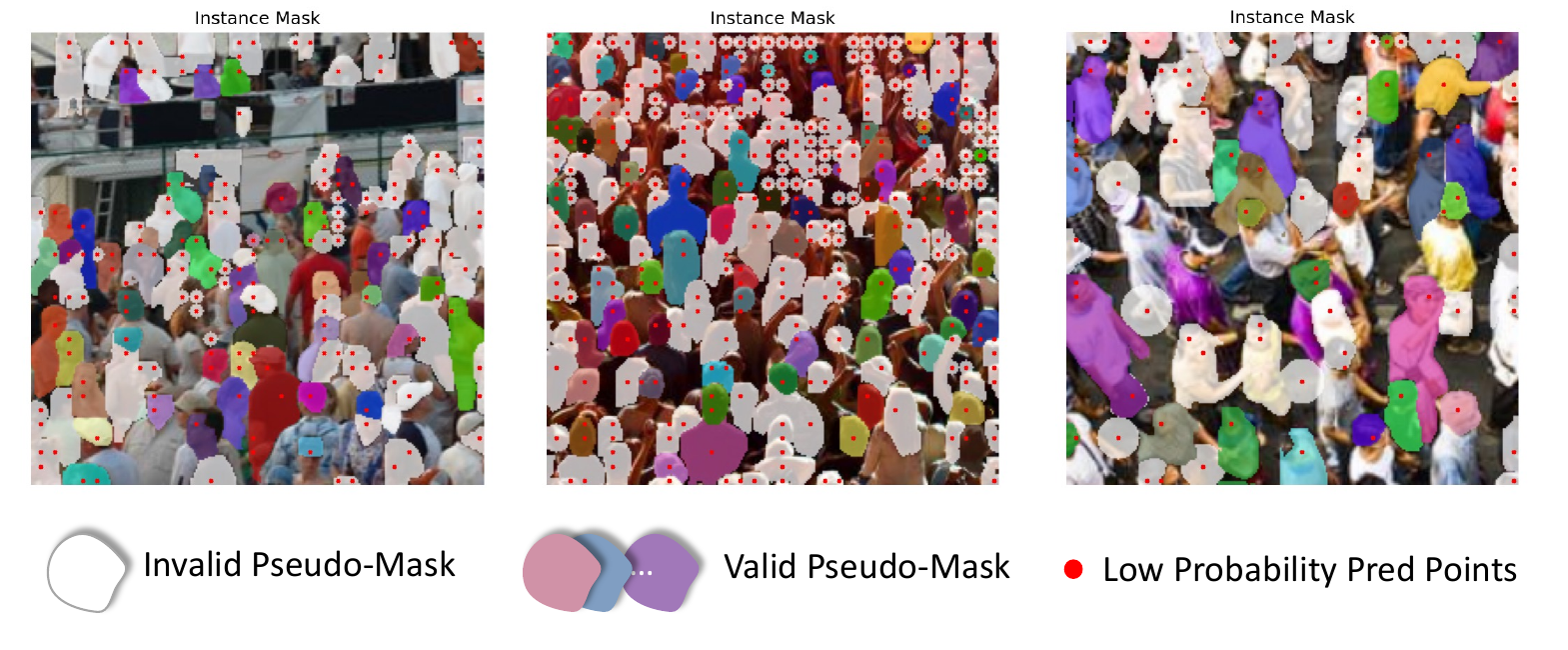}
    \caption{Pseudo-Mask Selection. We utilize a low threshold to obtain pseudo points, thereby generating all possible masks and ensuring the reliability of background. Then, based on the probability, we select the mask associated with the highest probability as the final valid mask. }
    \label{fig:Pseudo mask Vis}
    % \vspace{-5mm}
\end{figure}

\textbf{Adaptive Region-Aware Instantiation.} To reasonably assign computational regions to each instance, we employ adaptive radius calculation based on the NNEC between points. This approach automatically infers the local supervision range for each point according to the spatial distribution of annotated points. First, we calculate the Euclidean distance matrix between all annotated points in the sample. If there is exactly one annotated point, the radius degenerates into a global scale estimate. Radius for each point is defined as:

\begin{equation}
    % r_i^{(0)} =
    r_i =
    \begin{cases}
    \displaystyle \min_{j \neq i} \mathbf{D}_{ij}, 
    & N \ge 2 \\[6pt]
    \displaystyle \min(H, W)\cdot 0.5, 
    & N = 1
    \end{cases}, 
    \quad
    \mathbf{D}_{ij} = \lVert \mathbf{p}_i - \mathbf{p}_j \rVert_2.
    \label{nnec radius}
\end{equation}

% Subsequently, the radius undergoes range truncation to ensure it remains within a reasonable geometric scale:

% \begin{equation}
%     r_i = \mathrm{clamp}
%     \left(
%     r_i^{(0)},\ 
%     r_{\min},\ 
%     \min(H, W)\cdot \alpha_{\max}
%     \right),
%     \label{r clamp}
% \end{equation}

% here $r_{\min} = 0.5$ is the lower bound of the minimum radius, $\alpha_{\max} = 0.5$ is the max radius factor. 
We then define the spatial coordinate grid $G$, so the area contributing to the calculation for each final point is:

\begin{equation}
    C_i(y,x) = 
    \mathbf{1}
    \left[
    \lVert \mathbf{G}(y,x) - \mathbf{p}_i \rVert_2 \le r_i
    \right], 
    \quad
    \mathbf{G} \in \mathbb{R}^{H \times W \times 2}.
    \label{region candidate}
\end{equation}

Therefore, for each point, the corresponding GT or pseudo-mask is used to segment positive and negative sample regions within the candidate area $\mathcal{R}$:
\begin{equation}
    \mathcal{R}_i^{+}
    =
    C_i
    \cap
    \left\{
    (y,x)\mid \mathbf{M}(y,x) = \ell_i
    \right\},
    \quad
    \mathcal{R}_i^{-}
    =
    C_i
    \cap
    \left\{
    (y,x)\mid \mathbf{M}(y,x) \ne \ell_i
    \right\},
    \label{sample region}
\end{equation}
where $\ell_i$ is the instance ID of point $p_i$.

\textbf{Differentiable Center Sampling.} Next, to obtain the embedding vector at each annotation point as the instance prototype, we use differentiable grid sampling, and  $\bar{\mathbf{p}}_i$ is the normalized coordinate, placing the annotation point at the sub-pixel position. Thus, the formulation of the center feature is:

\begin{equation}
    \mathbf{c}_i = \mathrm{GridSample}(\tilde{\mathbf{E}}, \bar{\mathbf{p}}_i) 
    \in \mathbb{R}^{D},
    \quad
    \bar{\mathbf{p}}_i = \frac{2 \cdot \mathbf{p}_i}{(W-1,\, H-1)} - 1.
    \label{gridsample}
\end{equation}

Consequently, the final calculation of maskhead discriminative loss is combined with Pull force Loss and Push force Loss:

% \begin{equation}
%     \mathcal{L}_i^{+}
%     =
%     \frac{1}{\left|\mathcal{R}_i^{+}\right|}
%     \sum_{(y,x)\in \mathcal{R}_i^{+}}
%     \left[
%     d_i(y,x) - (\tau - \delta)
%     \right]_{+},
% \end{equation}
% \begin{equation}
%     \mathcal{L}_i^{-}
%     =
%     \frac{1}{\left|\mathcal{R}_i^{-}\right|}
%     \sum_{(y,x)\in \mathcal{R}_i^{-}}
%     \left[
%     (\tau + 4\delta) - d_i(y,x)
%     \right]_{+},
% \end{equation}

% \begin{equation}
%     d_i(y,x) 
%     =
%     \left\lVert 
%     \tilde{\mathbf{E}}_{\mathrm{flat}}(y,x) - \mathbf{c}_i
%     \right\rVert_2,
%     \label{Ls disatance}
% \end{equation}

\begin{equation}
\mathcal{L}_i
=
\frac{1}{|\mathcal{R}_i|}
\sum_{(y,x)\in \mathcal{R}_i}
\left[
\alpha_{y,x}
\left(
\left\lVert 
\tilde{\mathbf{E}}(y,x)-\mathbf{c}_i
\right\rVert_2
-(\tau - \alpha_{y,x} \cdot \delta)
\right)
\right]_+ ,
\end{equation}
where, $[x]_{+} = \max(x, 0)$ serves as the hinge function, and $\alpha_{y,x}\in\{+1,-1\}$ denotes the region label, with $+1$ for positive pixels and $-1$ for negative pixels. $\tau$ and $\delta$ are hyperparameters obtained from the experiment shown in supplement.
% The final loss is obtained by averaging across all valid points and normalizing along the batch dimension, here $M_b$ is the number of mask in one batch and \textit{B} is batch size: 
% \begin{equation}
%     L = \frac{1}{B} \sum_{b=1}^{B} 
%     \frac{1}{\lvert M_b \rvert} 
%     \sum_{i \in M_b} 
%     \left( L_i^{+} + L_i^{-} \right).
% \end{equation}

\subsubsection{Instance Segmentation Generation}
\label{sec:mask gen}
During the inference stage and pseudo-mask generation, given the feature map $\mathbf{E} \in \mathbb{R}^{D \times H \times W}$ output by the trained embedding network and the predicted human location point prompts $\mathcal{P} = \{\mathbf{p}_i\}_{i=1}^{N}$ , the overall generation of crowd segmentation is aligned with the aforementioned loss calculation. However, we observed that during the early stages of training, the generated pseudo-masks often exhibited incomplete clipping. To mitigate this issue, we propose incorporating additional spatial discriminative information by introducing a normalized squared geometric distance as a spatial regularization term. The computed embedding distance and geometric distance are fused into a joint energy function as follows:
\begin{equation}
    E_i(y,x) = 
    \left\lVert 
    \tilde{\mathbf{E}}(y,x)-\mathbf{c}_i
    \right\rVert_2 
    + \lambda \cdot \frac{\| G(y,x) - p_i \|_2^2}{(r_i + \epsilon)^2},
    \label{eq:energy}
\end{equation}
where $\lambda$ is the geometry weight and $\tilde{\mathbf{E}}$ is the smoothed feature map mentioned above. Defining $\tau_g$ as an energy threshold in the inference, the final segmentation S(y,x) of the instance for the entire image is as follows:
\begin{equation}
    \label{eq:final mask}
    S(y,x) =
    \begin{cases}
    i, & \text{if } E_i(y,x) < \tau_g \\
    0, & \text{otherwise}
    \end{cases},
\end{equation} where $i$ is the id number of each point. 

\subsection{Mask Constraint Counting}
\label{sec:counting and semi-framework}

Given that region masks still exhibit inaccuracies and lack semantic information, we propose incorporating mask constraints, utilizing the generated instance mask as a pseudo-mask. As illustrated in \cref{fig:3-stage-framework}, we introduce mask-constraint optimization as the last step to formulate the whole feedback framework. 

To mitigate teacher uncertainty in pseudo-mask generation, we perform instance segmentation on points with confidence above a low threshold, while only masks exceeding a higher threshold are retained as valid for loss computation. 
% Regions between the two thresholds are treated as invalid and excluded from both foreground and background supervision. 
This strategy improves the reliability of pseudo-mask learning, as visualized in \cref{fig:Pseudo mask Vis}. Based on this, we designed two computational methods as follows:

% As visualized in \cref{fig:3-stage-framework}, we purpose to add mask constraint optimizations after obtaining XMask model. Regarding pseudo mask generation we account for the inaccuracy of the teacher model by performing instance segmentation on points with a probability greater than low probability instead of 0.5. Masks with probabilities exceeding high probability are selected as the final valid masks for computation. Regions between low and high probability are treated as invalid mask areas, meaning they are excluded from both foreground and background calculations. This approach effectively ensures computational validity, and \cref{fig:Pseudo mask Vis} depicts our selection visualization.

\textbf{Background Penalty Loss.} The purpose of this loss function is to ensure that no response is present in the background region. As described in \cref{eq:bg loss}, gradients are induced only when the prediction for true background pixels is positive, driving them toward zero or negative values, and those that are already negative do not contribute additional loss. Here, $\Omega_0 = \{ p \mid M_{\mathrm{gt}}(p) = 0 \}$ denotes the background pixel geometry, and $\hat{f} \in \mathbb{R}^{H \times W}$ represents the bilinear interpolation output upscaled to the original image resolution.

\begin{equation}
    L_{\mathrm{bg}} =
    \frac{1}{\lvert \Omega_0 \rvert}
    \sum_{p \in \Omega_0}
    \max\big(0, \hat{f}(p)\big).
    \label{eq:bg loss}
\end{equation}

% \textbf{Foreground Instance Balance Loss.} The goal of this computation is to ensure consistent prediction probabilities within each mask, guaranteeing stable numerical outputs, mitigating significant amplitude fluctuations. First, sum the probabilities of points considered foreground within the mask region of each instance $k$, and apply a bilateral hinge loss to the integral value of each instance. As shown in \cref{eq:balance loss}, where $Kv$ denotes the set of valid instance IDs after pseudo-label filtering, and $\Omega_k = \{ p \mid M_{\mathrm{gt}}(p) = k \}$ represents the pixel set for the $kth$ instance. 

% \begin{equation}
%     \mathcal{L}_{\text{fg}} = \frac{1}{|\mathcal{K}_v|} \sum_{k \in \mathcal{K}_v} |S_k - 1|,   \quad
%     S_k = \sum_{p \in \Omega_k} \max\big(0, \hat{f}(p)\big)\
%     \label{eq:balance loss}
% \end{equation}

% \textbf{One-Point Loss.} Perform a one-hot encoding directly on each point predicted as foreground within each mask, with an expected value of 1. Utilize the one-hot matrix to count the number of positive response pixels within each instance mask in a single pass via matrix multiplication, and then apply a constraint with a target value of 1 to the positive pixel count for each instance to compute the loss:

% \begin{equation}
%     L_{\mathrm{one}} =
%     \frac{1}{\lvert K_v \rvert}
%     \sum_{k \in K_v}
%     \left| N_k^{+} - 1 \right|, 
%     \quad
%     N_k^{+} =
%     \sum_{p \in \Omega_k}
%     \mathbf{1}\big[\hat{f}(p) > 0\big].
%     \label{point loss}
% \end{equation}

\textbf{Foreground Constraint Loss.} To enforce instance-level consistency within each mask, we constrain both the aggregated foreground response and the number of activated pixels, with the one-point constraint serving as the primary objective. The foreground constraint loss is defined as:
\begin{equation}
\mathcal{L}_{\mathrm{fg}}
=
\frac{1}{|K_v|}
\sum_{k \in K_v}
\left(
\left| N_k^{+} - 1 \right|
+
\lambda
\left| S_k - 1 \right|
\right),
\end{equation}

\begin{equation}
S_k = \sum_{p \in \Omega_k} \max(0, \hat{f}(p)),
\quad
N_k^{+} = \sum_{p \in \Omega_k} \mathbf{1}[\hat{f}(p) > 0],
\end{equation}
where $K_v$ denotes the set of valid instance IDs after pseudo-label filtering, and $\Omega_k = \{ p \mid M_{\mathrm{gt}}(p) = k \}$ represents the pixel set for the $k$-th instance.

% Although large segmentation models such as SAM \cite{5-kirillov2023segment} and FastSAM \cite{49-zhao2023fastsegment} , their prohibitive inference cost in crowded scenes limits practicality. In contrast, our approach enables efficient end-to-end mask and pseudo-mask generation.

Prior to this, we also attempted to use existing segmentation large models (such as SAM \cite{5-kirillov2023segment} and FastSAM \cite{49-zhao2023fastsegment}) as a method for generating pseudo-masks. However, their prohibitive inference cost in crowded scenes limits practicality.

\section{Experiments}
We evaluate the proposed method for crowd instance segmentation and counting on three crowd counting datasets: ShTech A\cite{50-Zhang2016SingleImageCC}, UCF-QNRF\cite{51-idrees2018compositionlosscountingdensity}, and JHU++\cite{52-sindagi2020jhucrowdlargescalecrowdcounting}. Following P2R\cite{21-lin2025pointtoregionlosssemisupervisedpointbased}, DAC \cite{16-lin2022semisupervisedcrowdcountingdensity}, and OT-M \cite{22-Lin_2023_CVPR}, three protocols are applied: 5\%, 10\%, and 40\% of labeled data, with the remaining crowd images involved in training without annotations. The labeled samples in each percentage are chosen to align with DAC \cite{16-lin2022semisupervisedcrowdcountingdensity}. In ablation studies, we evaluated the impact of different architectures and hyperparameters to demonstrate the validity of our framework and the performance of our results.

\textbf{Settings}  For all three datasets, the random crop with a size of 256×256 is implemented, and we limit the shorter side of each image to 1536 and 2048 for UCF-QNRF\cite{51-idrees2018compositionlosscountingdensity} and JHU++\cite{52-sindagi2020jhucrowdlargescalecrowdcounting}, respectively while ShTech A\cite{50-Zhang2016SingleImageCC} is still the original size. For data augmentation, labeled data is directly input into the student model, while unlabeled data is weakly augmented for the teacher model and strongly augmented for the student model.

\subsection{Semi-supervised Crowd Instance Segmentation}

\begin{figure}[t]
    \centering
    \includegraphics[width=1.0\linewidth]{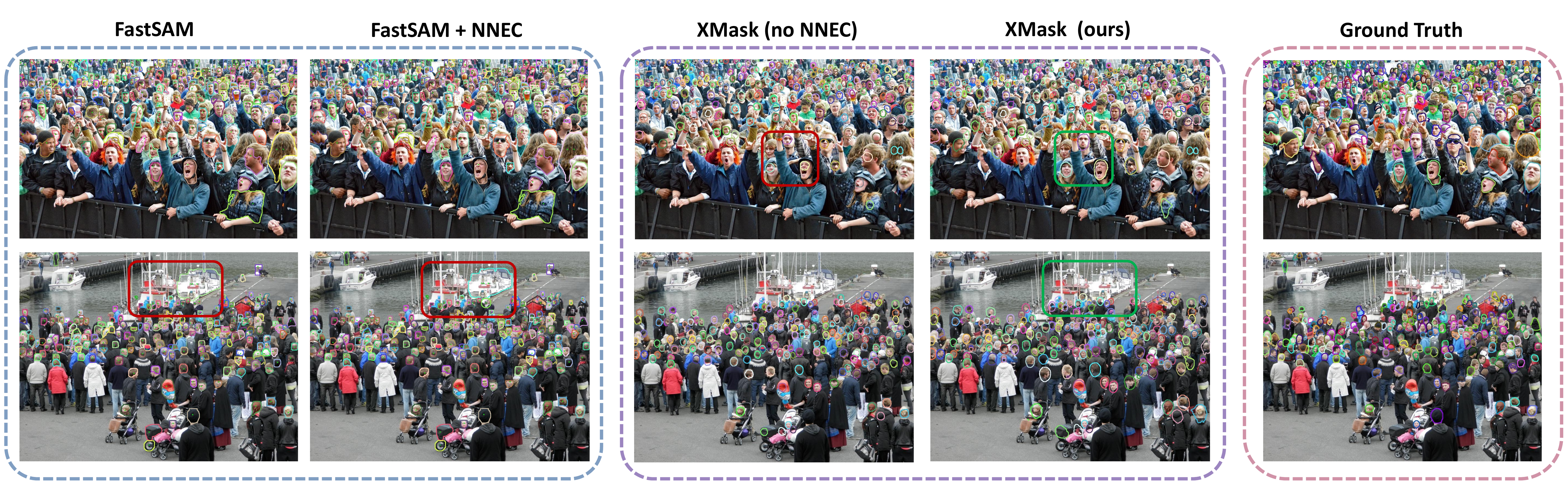}
    \caption{Visualization of masks generated by FastSAM-based, XMask, and ground truth generated by EDP-SAM. Through feature learning, our method demonstrates strong robustness in human segmentation, potentially segmenting other irrelevant objects compared to FastSAM. Furthermore, incorporating NNEC enhances the segmentation's dimensional adaptability.}
    \label{fig:Mask vis compare}
    \vspace{-5mm}
\end{figure}

This section demonstrates semi-supervised crowd instance segmentation performance in three datasets. As shown in \cref{fig:Mask vis compare}, our XMask approach is compared with three different point-prompt methods. All methods utilize identical point predictions from a unified semi-supervised counting model, ensuring consistent localization quality and isolating the segmentation module's contribution. Our approach demonstrates excellent performance in crowd segmentation tasks, achieving high recall masks across all test datasets. This is achieved through the utilization of predicted point constraints and the design of exclusivity feature learning methods.

% Regarding the baseline methods we used for comparison. Initially, FastSAM reformulates the original SAM encoder–decoder pipeline into a real-time, single-stage CNN framework based on YOLOv8-seg to generate class-agnostic candidate masks, followed by prompt-guided post-processing for target retrieval. Given the set of candidate masks from FastSAM\cite{49-zhao2023fastsegment} everything mode and a set of point prompts, \textbf{FastSAM\cite{49-zhao2023fastsegment} only}  method assigns each point to the smallest-area mask that covers it via vectorized hit-matrix indexing, while \textbf{FastSAM\cite{49-zhao2023fastsegment} + NNEC} strategy adopts a conditional masking mechanism: each point first queries candidate masks under the smallest-area hit criterion; if a valid mask exists, it is directly assigned to the instance, otherwise the point falls back to an NNEC-generated circle as a geometric proxy.

As a baseline, FastSAM \cite{49-zhao2023fastsegment} reformulates SAM into a real-time framework to generate class-agnostic masks with prompt-guided refinement. The FastSAM-only variant assigns each point to the smallest candidate mask covering it. In contrast, FastSAM + NNEC introduces a conditional fallback strategy, using an NNEC-generated circular proxy when no valid mask is found.

As presented in \cref{tab:Seg test}, we evaluate IoU@0.5 with segmentation generated by EDP-SAM. Our XMask method consistently achieves excellent performance with substantial improvements over baselines. Ablation between our method variants reveals the complementary role of the geometric prior: on denser datasets (UCF-QNRF\cite{51-idrees2018compositionlosscountingdensity} and JHU++\cite{52-sindagi2020jhucrowdlargescalecrowdcounting}), the NNEC-augmented version yields consistent improvements, validating the conditional circle fallback for points lacking valid mask coverage. XMask also demonstrates significant computational efficiency, averaging 0.17s per image on ShTech A compared to 1.01s for FastSAM\cite{49-zhao2023fastsegment} (5.9× speedup). More efficiency comparisons are demonstrated in \cref{sec:ablation}.

To further strengthen the reliability of results, we conduct comprehensive hyperparameter sensitivity analyses on XMask ($\lambda$ and $\tau_g$ in \cref{eq:energy,eq:final mask}) and FastSAM\cite{49-zhao2023fastsegment} (IoU and conf), which are shown in the supplementary material. Additionally, for possible circularity concern, we clarify that both baseline CrowdSAM\cite{39-cai2024crowdsamsamsmartannotator} and FastSAM\cite{49-zhao2023fastsegment} employ the same
NNEC-based constraint for fair comparison and the mask annotations were refined through manual
correction (e.g., dense regions, boundary errors) to mitigate potential bias
% , which are shown in \cref{FastSAm and SS-RDCS Mask generation hyperparameters}.

\begin{figure}[!t]
    \centering
    \includegraphics[width=1.0\linewidth]{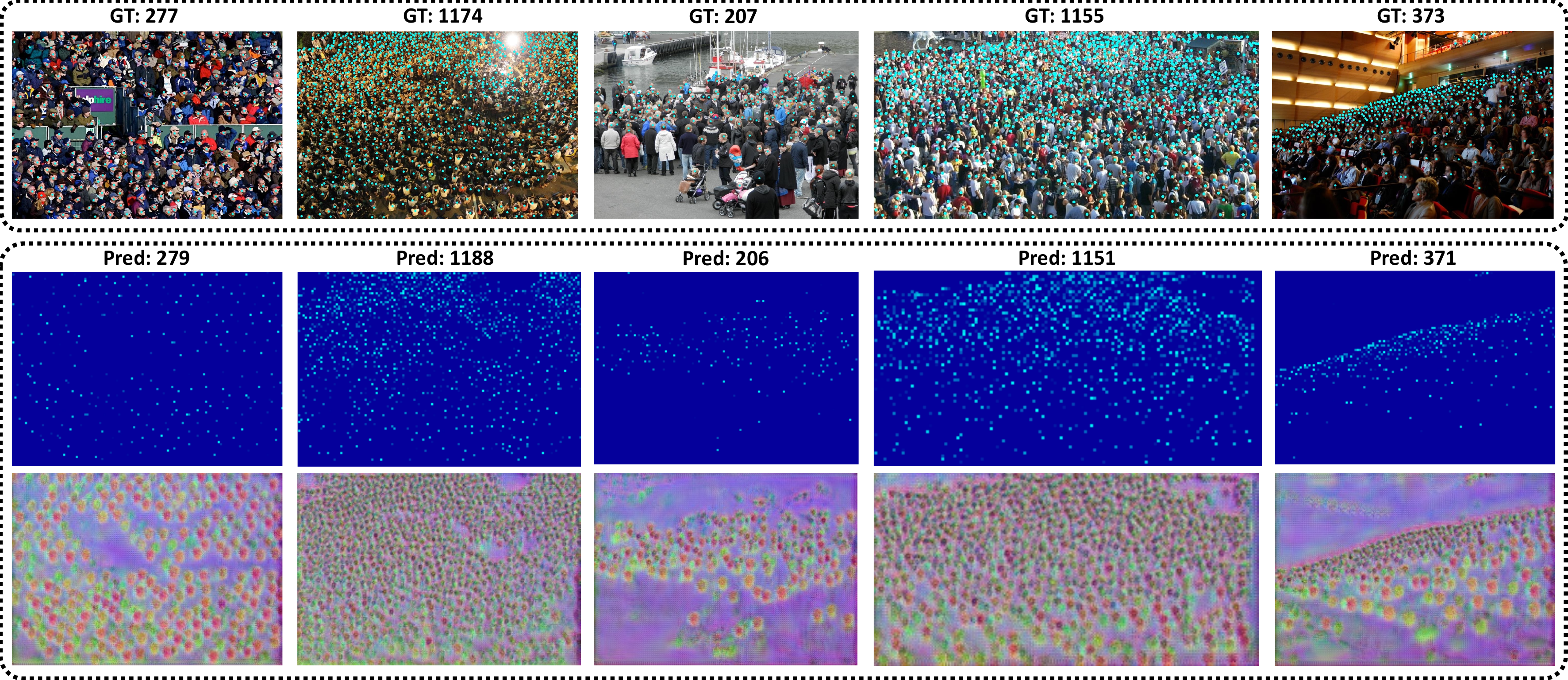}
    \caption{Visualization of Counting results on ShTech A. The first row: GT location. The second row: predicted points. The third row: feature map from maskhead in our method. In both dense and normal crowd scenes, our method yields counting predictions that closely align with the ground-truth counts, while the learned feature maps maintain structural coherence with the semantic layout of crowd head regions.}
    \label{fig:counting visualization}
    % \vspace{-5mm}
\end{figure}

\begin{table}[tb]
\caption{Semi-supervised crowd instance segmentation comparison on IoU@0.5. }
\label{tab:Seg test}
\centering
\scriptsize              
\setlength{\tabcolsep}{2pt} 
\begin{tabular}{c|c|ccc|ccc|ccc}
\toprule
\multirow{2}*{\parbox[c]{2.0cm}{\centering Methods}}
& \multirow{2}*{\parbox[c]{1.0cm}{\centering NNEC}}
& \multicolumn{3}{c|}{ShTech A\cite{50-Zhang2016SingleImageCC}} 
& \multicolumn{3}{c|}{UCF-QNRF\cite{51-idrees2018compositionlosscountingdensity}} 
& \multicolumn{3}{c}{JHU++\cite{52-sindagi2020jhucrowdlargescalecrowdcounting}} \\
&&5\% & 10\% & 40\% & 5\%  & 10\% & 40\% & 5\%  & 10\% & 40\% \\
\midrule
FastSAM\cite{49-zhao2023fastsegment} & & 0.279 & 0.281 & 0.291 & 0.203 & 0.211 & 0.213 & 0.186 & 0.185 & 0.189\\
FastSAM\cite{49-zhao2023fastsegment} & \checkmark & 0.287 & 0.290 & 0.300 & 0.223 & 0.229 & 0.233 & 0.197 & 0.182 & 0.201\\

S4M \cite{6-yoon2025s4mboostingsemisupervisedinstance}
& & 0.005 & 0.007 & 0.011 
& 0.010 & 0.012 & 0.015 
& 0.013 & 0.014 & 0.021 \\

CrowdSAM \cite{39-cai2024crowdsamsamsmartannotator}
& \checkmark & 0.201 & 0.201 & 0.212 
& 0.149 & 0.156 & 0.163 
& 0.187 & 0.189 & 0.192 \\

point only& \checkmark &0.195 & 0.194 & 0.202 & 0.198 & 0.201 & 0.207 & 0.180 & 0.182 & 0.183 \\
\midrule
XMask(ours) & &\textbf{0.314} & \textbf{0.318} & \underline{0.331} & \underline{0.269} & \underline{0.277} & \underline{0.278} & \underline{0.249} & \underline{0.253} & \underline{0.257} \\
XMask (ours) & \checkmark &\underline{0.313} & \textbf{0.318} & \textbf{0.333} & \textbf{0.282} & \textbf{0.290} & \textbf{0.297} & \textbf{0.258} & \textbf{0.263} & \textbf{0.270} \\
\bottomrule
\end{tabular}
\vspace{-5mm}
\end{table}

\begin{table}[tb]
\caption{Semi-supervised crowd counting performance comparison between recent methods and our framework on three datasets under various labeled protocols.}
\label{tab:counting test}
\centering
\scriptsize
\setlength{\tabcolsep}{4.0pt}
\renewcommand{\arraystretch}{1.0}
\newcommand{\blankimp}[1]{\phantom{\textcolor{blue}{\tiny$\downarrow$#1}}}
\newcommand{\down}[1]{\textcolor{blue}{\tiny$\downarrow$#1}}
    \begin{tabular}{c|c|cc|cc|cc}
    \toprule
    
    \multirow{2}*{\parbox[c]{1.5cm}{\centering labeled Pct.}}
    & \multirow{2}*{\parbox[c]{1.5cm}{\centering Methods}}
    & \multicolumn{2}{c|}{ShTech A~\cite{50-Zhang2016SingleImageCC}}
    & \multicolumn{2}{c|}{UCF-QNRF~\cite{51-idrees2018compositionlosscountingdensity}}
    & \multicolumn{2}{c}{JHU++~\cite{52-sindagi2020jhucrowdlargescalecrowdcounting}} \\
    
    & 
    & {MAE↓} & {MSE↓}
    & {MAE↓} & {MSE↓}
    & {MAE↓} & {MSE↓} \\
    
    \midrule
    % \hline\hline
    
    \multirow{6}*{\parbox[c]{1.0cm}{\centering 5\%}}
    & MT~\cite{72-springenberg2015strivingsimplicityconvolutionalnet} 
    & 104.7 & 33.2 & 172.4 & 284.9 & 101.5 & 363.5 \\
    & L2R~\cite{14-liu2018leveragingunlabeleddatacrowd}
    & 103.0 & 27.6 & 160.1 & 272.3 & 101.4 & 338.8 \\
    & DAC~\cite{16-lin2022semisupervisedcrowdcountingdensity}
    & 85.4 & 134.5 & 120.2 & 209.3 & 82.2 & 294.9 \\
    & OT-M~\cite{22-Lin_2023_CVPR}
    & 83.7 & 133.3 & 118.4 & 195.4 & 82.7 & 304.5 \\
    % & P³Net~\cite{17-lin2024semisupervisedcountingpixelbypixeldensity}
    % & 85.5 & 131.0 & 115.3 & 195.2 & 80.8 & 306.1 \\
    & P2R~\cite{21-lin2025pointtoregionlosssemisupervisedpointbased} 
    & 69.9 
    & 119.5 
    & 100.1 
    & 182.5 
    & 77.8 
    & 293.5 \\
    
    & Ours
    & {\textbf{64.6}} 
    & {\textbf{106.2}} 
    & {\textbf{98.4}} 
    & {\textbf{163.7}} 
    & {\textbf{77.1}} 
    & {\textbf{289.5}} \\
    
    \midrule
    
    \multirow{6}*{\parbox[c|]{1.0cm}{\centering 10\%}}

    & MT~\cite{72-springenberg2015strivingsimplicityconvolutionalnet} 
    & 319.3 & 94.5 & 156.1 & 245.5 & 250.3 & 90.2 \\
    & L2R~\cite{14-liu2018leveragingunlabeleddatacrowd}
    & 90.3 & 153.5 & 148.9 & 249.8 & 87.5 & 315.3 \\
    & DAC~\cite{16-lin2022semisupervisedcrowdcountingdensity}
    & 74.9 & 115.5 & 109.0 & 187.2 & 75.9 & 282.3 \\
    & OT-M~\cite{22-Lin_2023_CVPR}
    & 80.1 & 118.5 & 113.1 & 186.7 & 73.0 & 280.6 \\    
    % & P³Net~\cite{17-lin2024semisupervisedcountingpixelbypixeldensity}
    % & 72.1 & 116.4 &103.4 & 179.0 & 71.8 & 294.4 \\
    
    & P2R~\cite{21-lin2025pointtoregionlosssemisupervisedpointbased} 
    & 64.2 
    & 114.6 
    & 94.9 
    & 167.2 
    & \textbf{68.7} 
    & 272.3 \\
    
    & Ours
    & {\textbf{62.4}} 
    & {\textbf{104.4}} 
    & {\textbf{93.9}} 
    & {\textbf{155.7}} 
    & {\textbf{68.7}} 
    & {\textbf{258.7}}\\
    
    \midrule
    
    \multirow{6}*{\parbox[c|]{1.0cm}{\centering 40\%}}

    & MT~\cite{72-springenberg2015strivingsimplicityconvolutionalnet} 
    & 88.2 & 151.1 & 147.2 & 249.6 & 121.5 & 388.9 \\
    & L2R~\cite{14-liu2018leveragingunlabeleddatacrowd}
    & 86.5 & 148.2 & 145.1 & 256.1 & 123.6 & 376.1 \\
    & DAC~\cite{16-lin2022semisupervisedcrowdcountingdensity}
    & 67.5 & 110.7 & 91.1 & 153.4 & 65.1 & 260.0 \\
    & OT-M~\cite{22-Lin_2023_CVPR}
    & 70.7 & 114.5 & 100.6 & 167.6 & 72.1 & 272.0 \\
    % & P³Net~\cite{17-lin2024semisupervisedcountingpixelbypixeldensity}
    % & 63.0 & 100.9 & 90.0 & 155.4 & 58.9 & \textbf{251.9} \\
    
    & P2R~\cite{21-lin2025pointtoregionlosssemisupervisedpointbased} 
    & \textbf{55.6} 
    & 95.0 
    & 86.0 
    & 144.3 
    & 63.3 
    & 271.1 \\
    
    & Ours
    & 56.2 
    & {\textbf{93.5}} 
    & {\textbf{84.0}} 
    & {\textbf{143.6}} 
    & {\textbf{61.9}} 
    & {\textbf{256.1}} \\
    
    \bottomrule
    \end{tabular}
    \vspace{-5mm}
\end{table}

\subsection{Semi-Supervised Crowd Counting}

We further evaluate the crowd counting accuracy against the previous state-of-the-art methods. \cref{tab:counting test} shows that our framework consistently outperforms across nearly all three benchmarks, with improvements particularly pronounced under limited annotated images. By leveraging reliable pseudo-masks and structural consistency learning from XMask, our method outperforms P2R \cite{21-lin2025pointtoregionlosssemisupervisedpointbased}, the base counting model used in Stage 1. As visualized in \cref{fig:counting visualization}, our approach delivers accurate counting predictions while enforcing structural coherence between features and head-region semantics.

Under the 5\% setting, our proposed framework achieves the best performance in nearly all protocols across all crowd datasets, demonstrating that the learned segmentation priors effectively compensate for limited supervision. As label ratios increase to 10\% and 40\%, our method maintains consistent gains, with notably substantial MSE improvements, indicating enhanced robustness against large estimation outliers. 
% These results validate both the effectiveness of our counting-enhanced pipeline and the rationality of the mask loss design. 

\subsection{Ablation study}
\label{sec:ablation}

% \jia{Any ablation about EDP-SAM and XMask? e.g. EDP-SAM compared with SAM, XMask compared with normal segmentation loss?}
% \begin{figure}

We then ablate the components in the proposed method and analyze the hyperparameters.
% This section presents ablation experiments, including relative training hyperparameters and components. % More tests will be shown in supplements.
% includes experiments on hyperparameters threshold $\tau$ used in mask discriminative loss, the Depthwise Gaussian Smoothing and Differentiable Sampling components in XMask, and the selection of training defrosting structures for the final stage. 

\begin{wrapfigure}{r}{0.5\linewidth}
    \centering
    \vspace{-5mm}
    \includegraphics[width=0.95\linewidth]{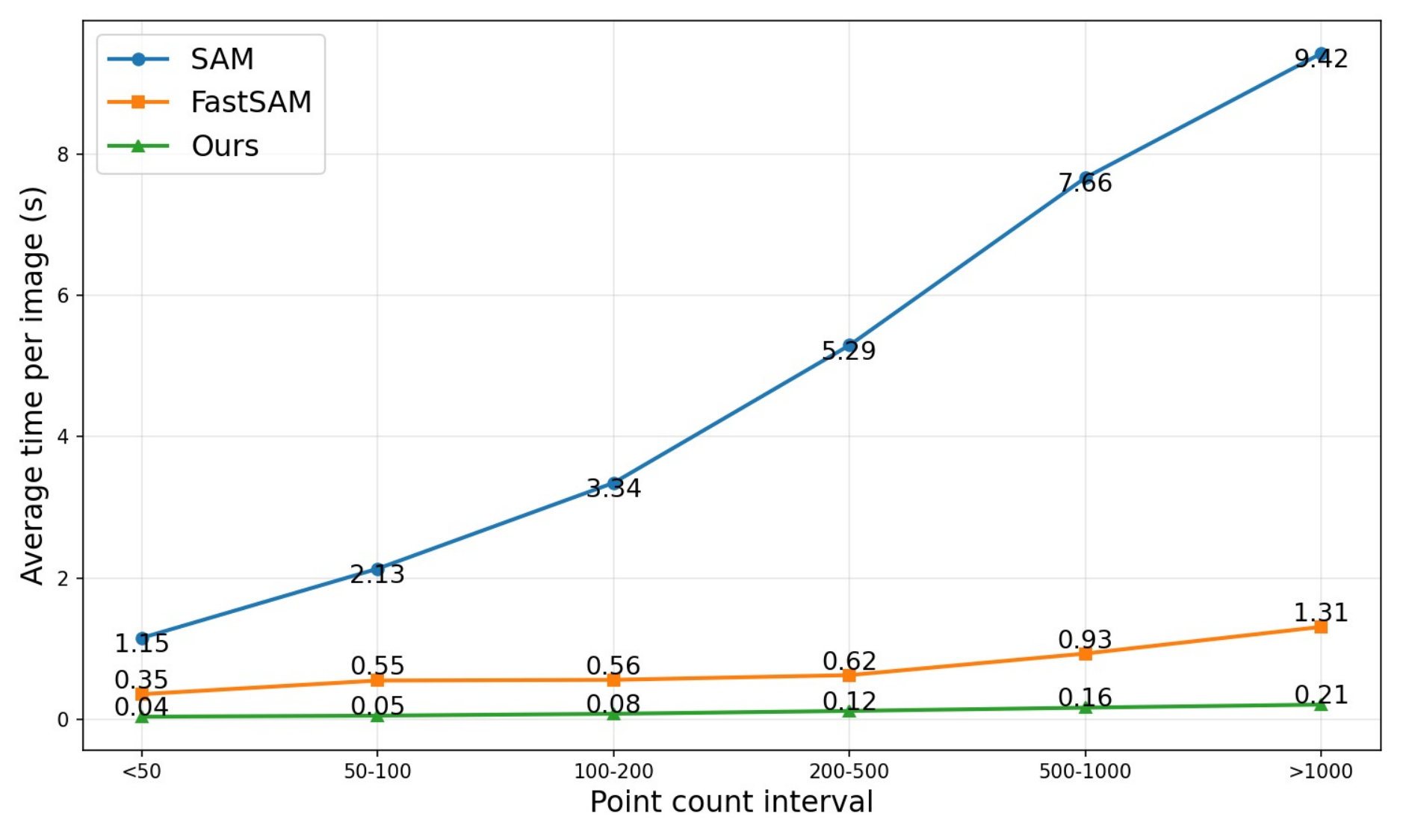}
    \caption{Our method consistently outperforms SAM and FastSAM in segmentation efficiency across varying crowd densities, maintaining shorter processing times despite differences in density.}
    \label{fig:efficiency}
    \vspace{-6mm}
% \end{figure}
\end{wrapfigure}

\textbf{Segmentation Efficiency.} We conducted segmentation inference experiments on the ShTechA test dataset, which was divided into multiple density intervals for comparison. As shown in \cref{fig:efficiency}, compared to FastSAM and SAM, our method significantly improves segmentation speed, achieving less than 0.3 seconds per image across all density intervals, while demonstrating insensitivity to density. Therefore, our approach is highly advantageous for training efficiency when generating pseudo masks in semi-supervised settings.

\textbf{Loss Function Comparison.} In semi-supervised segmentation learning, we propose utilizing the maskhead discriminative loss. We compared it with several other segmentation losses on the ShTechA dataset with 5\% labeled data, including Dice loss and Focal loss. Results in \cref{subtab:loss} demonstrate that our approach achieves the best performance in terms of mask IoU and F1 score.

\textbf{Differentiable Center Sampling.}  
Experiments on ShTech A (\cref{subtab:GridSample Test}) show that feature sampling slightly improves performance. Despite the modest gain, we adopt differentiable sampling over discrete extraction due to its sub-pixel precision and smoother gradient propagation, which lead to more stable and optimization-friendly feature embeddings.
% Experiments on the ShTech A dataset \cite{50-Zhang2016SingleImageCC} (\cref{subtab:GridSample Test}) demonstrate the benefits of using Sampling for feature capture. Although the performance gain is subtle, we prioritize Differentiable Sampling over discrete extraction to leverage its differentiability. This ensures sub-pixel precision and smoother gradient propagation, resulting in more robust feature embeddings suitable for optimization.

\textbf{Depthwise Gaussian Smoothing.} 
This module applies Gaussian smoothing to feature embeddings before pairwise L2 distance computation, enhancing spatial coherence and reducing noisy activations. \cref{subtab:Gaussian test} shows that Gaussian filtering consistently improves segmentation performance and a kernel size of 7 with $\sigma=3$ achieves the best results. % Among different configurations, 
% This module serves as a preprocessing step prior to computing the pairwise L2 distance in the feature space. By applying Gaussian filtering to the extracted feature embeddings, this operation promotes spatial coherence, suppresses isolated noisy activations, and yields smoother and more spatially consistent segmentation masks. To investigate its effect, we conduct an ablation study on the ShTech A dataset under the 5\% labeled data setting. As reported in \cref{subtab:Gaussian test}, incorporating Gaussian smoothing into the feature embeddings leads to consistent improvements in segmentation performance over the baseline without smoothing. We further examine the sensitivity to the kernel size and standard deviation $\sigma$, and observe that the configuration of kernel size = 7 and $\sigma$ = 3 achieves the best performance. 

\textbf{Trained Modules.} 
In the final counting stage, we explore different fine-tuning strategies to balance performance and efficiency. As shown in \cref{subtab:Stage3 trained modules}, unfreezing only the decoder achieves the best counting accuracy. Therefore, we adopt this setting as the default configuration. % for Stage 3.

\begin{table*}[t]
\centering
\tiny
\setlength{\tabcolsep}{3pt}
\renewcommand{\arraystretch}{0.92}
\caption{Comprehensive ablation studies on different components and settings.}
\vspace{-3mm}
\label{tab:ablation_all}
\begin{subtable}[t]{0.52\linewidth}
\centering
\scriptsize
\caption{Loss function comparison on ShTech A. Dice loss calculates overlap-based ratio between prediction and ground truth, while Focal loss weighted cross-entropy focusing on hard samples.}
\label{subtab:loss}
\tiny
\begin{tabular}{c|cc}
\toprule
Loss & IoU↑ & F1↑ \\
\midrule
Dice loss & 0.2381 & 0.0946 \\
Focal loss & 0.2893 & 0.2070\\
Ours  & \textbf{0.3119} & \textbf{0.2574} \\
%Dice + Ours & 0.3027 & 0.2394 \\
%Focal + Ours & 0.3052 & 0.2457\\
\bottomrule
\end{tabular}
\end{subtable}
\hfill
%--------------------------------------------
\begin{subtable}[t]{0.44\linewidth}
\centering
\caption{Gaussian Smoothing on ShTech A}
\label{subtab:Gaussian test}
\begin{tabular}{c|cc|cc}
\toprule
Gaussian & K & $\sigma$ & IoU↑ & F1↑ \\
\midrule
 & / & / & 0.3050 & 0.2481 \\
\checkmark & 7 & 1.5 & 0.3119 & 0.2574 \\
\checkmark & 5 & 1.5 & 0.3111 & 0.2566 \\
\checkmark & 9 & 1.5 & 0.3120 & 0.2577 \\
\checkmark & 7 & 3 & \textbf{0.3130} & \textbf{0.2584} \\
\checkmark & 5 & 3 & 0.3119 & 0.2575 \\
\checkmark & 9 & 3 & 0.3130 & 0.2570 \\
\bottomrule
\end{tabular}
\end{subtable}

\par\medskip

\begin{subtable}[t]{0.48\linewidth}
\centering
\caption{Differentiable Center Sampling on UCF}
\label{subtab:GridSample Test}
\begin{tabular}{c|cc|cc|cc}
\toprule
\multirow{2}*{\parbox[c]{0.8cm}{\centering Sample}}
& \multicolumn{2}{c|}{5\%}
& \multicolumn{2}{c|}{10\%}
& \multicolumn{2}{c}{40\%} \\
& IoU↑ & F1↑
& IoU↑ & F1↑
& IoU↑ & F1↑ \\
\midrule
& 0.2806 & 0.1973
& 0.2888 & 0.2111
& 0.2955 & 0.2079 \\

\checkmark
& \textbf{0.2807} & \textbf{0.1976}
& \textbf{0.2892} & \textbf{0.2129}
& \textbf{0.2956} & \textbf{0.2079} \\
\bottomrule
\end{tabular}
\end{subtable}
\hfill
%--------------------------------------------
\begin{subtable}[t]{0.48\linewidth}
\centering
\caption{Optimization Modules on ShTech A}
\label{subtab:Stage3 trained modules}
\begin{tabular}{cc|c|cc}
\toprule
Head & Backbone & Params & MAE↓ & MSE↓ \\
\midrule
\checkmark & \checkmark & 16.77M & 68.1 & 106.8 \\
\checkmark &  & 15.58M & 70.2 & 111.3 \\
 & \checkmark & 2.05M & \textbf{64.6} & \textbf{106.2} \\
\bottomrule
\vspace{-5mm}
\end{tabular}
\end{subtable}
\end{table*}

\section{Conclusion}
\label{conclusion}
In this work, we propose to supervise the semi-supervised crowd instance segmentation and counting based on mask supervision, which provides more structural individual shape information than point annotation. 
First, we propose an Exclusion-Constrained Dual-Prompt SAM (EDP-SAM) to generate robust masks from point annotations. Next, we combine EDP-SAM with manual correction to generate accurate mask annotations for the existing datasets.  To better segment instances in crowded images, we propose Exclusivity-Guided Mask Learning (XMask) that enforces spatial separation through a discriminative mask objective. 
% a semi-supervised crowd instance segmentation and counting framework based a  crowd-oriented EDP-SAM to generate mask supervision, an XMask learning strategy, and develop an efficient framework under a semi-supervised paradigm. 
Extensive experiments demonstrate that mask supervision effectively improves semi-supervised crowd instance segmentation and counting performance, offering a scalable solution to annotation scarcity. 
Future work will focus on improving boundary precision and refining point-generation strategies to achieve finer-grained segmentation quality.

% In this work, we propose a crowd-oriented EDP-SAM mask auto-generation method to alleviate the long-standing scarcity of semantic mask annotations in popular crowd counting datasets. Building upon this, we introduce an efficient and memory-friendly XMask network that explicitly bridges point supervision and instance segmentation under a unified semi-supervised learning paradigm for dense crowds. More importantly, by leveraging these high-quality pseudo-masks to regularize the counting objective, our Segmentation and Counting Feedback framewrok, improve the robust and accuracy in semi-supervised crowd counting.

% By jointly optimizing segmentation and counting within a single end-to-end framework, our method successfully bridges two traditionally separated tasks, achieving outstanding performance in both semi-supervised settings, demonstrating that pixel-level semantic recovery from point annotations can effectively enhance numerical estimation. This work provides a scalable and practical solution to the annotation bottleneck in crowd analysis, establishing a new paradigm for unified semi-supervised learning in dense scenes.

% While the proposed high-recall strategy proves highly effective for counting-oriented supervision, future work will focus on refining boundary precision and optimizing point generation strategies to further improve fine-grained segmentation quality.

\section*{Acknowledgments}
\label{Acknowledgments}
This work was supported by the National Natural Science Foundation of China under Project 62406090 and Jiangsu Science and Technology Major Program (Grant No. BG2024041).

% Now we have reached the maximum length of an ECCV \ECCVyear{} submission (excluding references and acknowledgements).
% References should start immediately after the main text, but can continue past p.\ 14 if needed. 
% \clearpage  % TODO FINAL: This \clearpage needs to be removed from both review and camera-ready versions.

% \section*{Acknowledgements}
% Please insert your acknowledgments here.

\bibliographystyle{splncs04}
\bibliography{main}

\clearpage
\appendix
\setcounter{figure}{0}
\setcounter{table}{0}
\setcounter{equation}{0}

\renewcommand{\thefigure}{S\arabic{figure}}
\renewcommand{\thetable}{S\arabic{table}}
\renewcommand{\theequation}{S\arabic{equation}}

\section{Implementation Details}

\textbf{Data pre-processing.} During training, all images are cropped to a resolution of
256 × 256. For labeled samples, each cropped image is horizontally flipped with a probability of 0.5 and randomly rescaled using a factor ranging from 0.7 to 1.3. For unlabeled samples, weak augmentations follow the same strategy as labeled data, while strong augmentations include color jittering (brightness, contrast, saturation, hue), grayscale conversion, Gaussian blur, and Cutout.

\textbf{training process.} In all experiments, we train the model for 1500 epochs with a batch size of 16. Only labeled data are used in the first 50 epochs for initialization in the first and second steps, and in the mask constraint optimization step, we do not initialize the training. Adam serves as the optimizer, with a learning rate of 5e-5 for decoder (point prediction head and mask head), 1e-5 for backbone in the first and second steps, while in the last step, the learning rate is set to various values depending on the specific dataset and labeled percentage. Additionally, in the mask constraint stage, we set the low probability threshold to 0.1 and the high threshold to 0.95 to guarantee the validation of the pseudo constraint.

\subsection{XMask Training Settings}

\textbf{Hyperparameters in MaskHead Discriminative Loss.} 
For the hyperparameters of the mask discriminative loss shown above, we tested training mask configurations with different label percentages on the ShTech A dataset using pre-trained counting model weights. Based on the ground-truth mask, we computed the L2 distance between the embeddings produced by the maskhead for regions inside and outside the mask, using the existing weights. As shown in \cref{tab:threshold in loss comptue}, the L2 feature distance between within and between instances is around 0.6. There, our XMask training threshold is 0.6, and the margin is 0.1.

\begin{figure}
    \centering
    \includegraphics[width=1.0\linewidth]{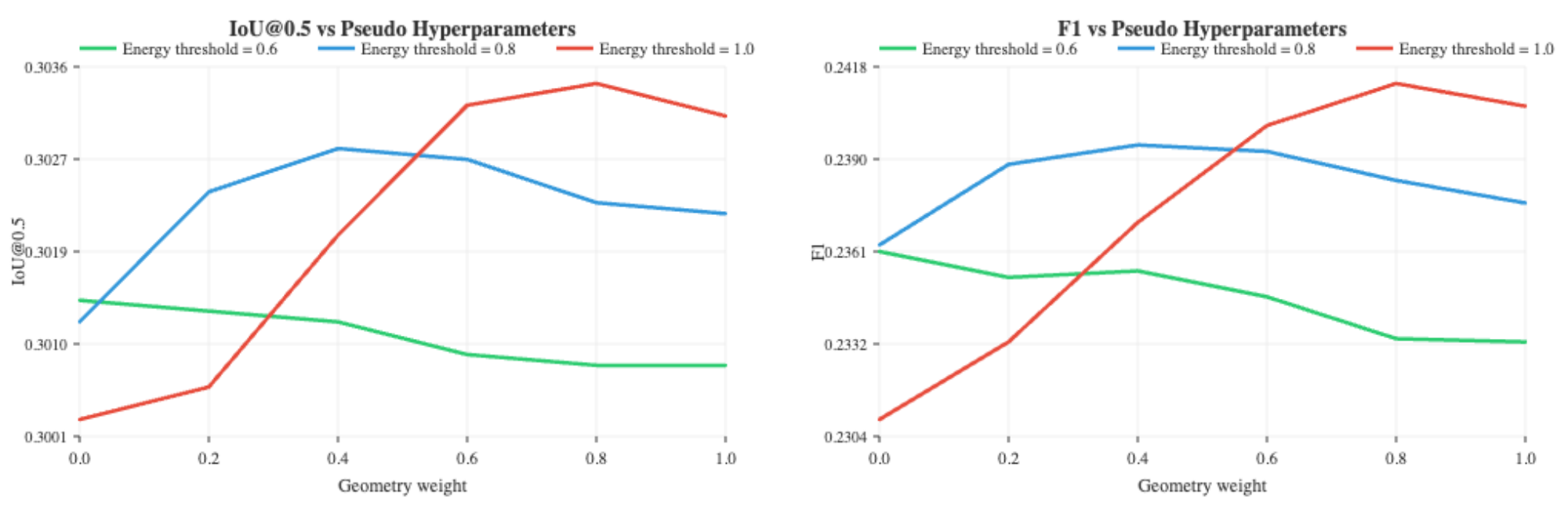}
    \caption{Experiment hyperparameters of Pseudo Mask generation in the training of XMask on ShTech A. We evaluated IoU@0.5 and F1 of XMask with different settings of $\lambda$ and $\tau_g$, and choose 1.0 and 0.8 respectively in the final XMask training.}
    \label{fig:pseudo mask test}
\end{figure}

\textbf{Hyperparameters in Pseudo Masks Generation.} Aligning with description in \cref{sec:seg inference}, there are two parameters $\lambda$ and $\tau_g$ in the generation of pseudo masks. According to \cref{fig:pseudo mask test}, we set $\lambda$ = 1.0 and $\tau_g$ = 0.8 in XMask training, and the trained model achieved the highest IoU@0.5 and F1.

\subsection{Segmentation Inference Settings}
\label{sec:seg inference}
In order to strengthen the reliability and equality of our segmentation comparison, we conducted experiments on hyperparameters in both FastSAM and our approach to achieve the best performance with high-recall segmentation.

\begin{table}[tb]
\centering
\setlength{\tabcolsep}{18pt}
\renewcommand{\arraystretch}{1.0}
\caption{ Loss threshold test on ShTech A. Before XMask training, we tested the initial L2 feature distance on the untrained feature map with ground truth masks.}
\label{tab:threshold in loss comptue}
\begin{tabular}{c|cc}
\toprule
\multirow{2}*{\parbox[c]{1.0cm}{\centering Labeled-Pct.}}
% & \multicolumn{2}{c|}{Test 1} 
% & \multicolumn{2}{c|}{Test 2} 
% & \multicolumn{2}{c|}{Test 3} 
& \multicolumn{2}{c}{Average L2 Feature Distance} \\
% & Intra & Inter 
% & Intra & Inter 
% & Intra & Inter 
& within same instance & between different instances \\
\midrule
% 40\% & 0.622 & 0.568 & 0.631 & 0.546 & 0.627 & 0.566 &
40\% & 0.630 & 0.560 \\
% 10\% & 0.589 & 0.598 & 0.606 & 0.549 & 0.607 & 0.5\cite{64-debrabandere2017semanticinstancesegmentationdiscriminative} &
10\% & 0.600 & 0.570 \\
% 5\%  & 0.591 & 0.571 & 0.635 & 0.526 & 0.607 & 0.553 & 
5\% & 0.610 & 0.550 \\
\bottomrule
\end{tabular}
\end{table}

\textbf{XMask(ours).}The generation of each mask S(y,x) matched with corresponding point relates two parameters $\tau_g$ and $\lambda$:
\begin{equation}
    E_i(y,x) = \tilde{\mathbf{E}} + \lambda \cdot \frac{\| G(y,x) - p_i \|_2^2}{(r_i + \epsilon)^2},
    \label{eq:energy}
\end{equation}

\begin{equation}
    \label{eq:final mask}
    S(y,x) =
    \begin{cases}
    i, & \text{if } E_i(y,x) < \tau_g \\
    0, & \text{otherwise}
    \end{cases},
\end{equation} where $i$ is the id number of each point, $E_i(y,x)$ is the energy and $G(y,x)$ is the position of pixels in circles matched with each points, $\tilde{E}$ is the feature distance with center feature, $p_i$ presents the center points location and $r_i$ is radius of circle of each points, and here circles is generated by NNEC method. 

\begin{figure}
\caption{Experiments of Hyperparameters used in Segmentation Task on ShTech A}
\label{FastSAm and SS-RDCS Mask generation hyperparameters}
\begin{subfigure}{1.0\linewidth}
    \label{fig:fastsam iou}
    \centering
    \includegraphics[width=\linewidth]{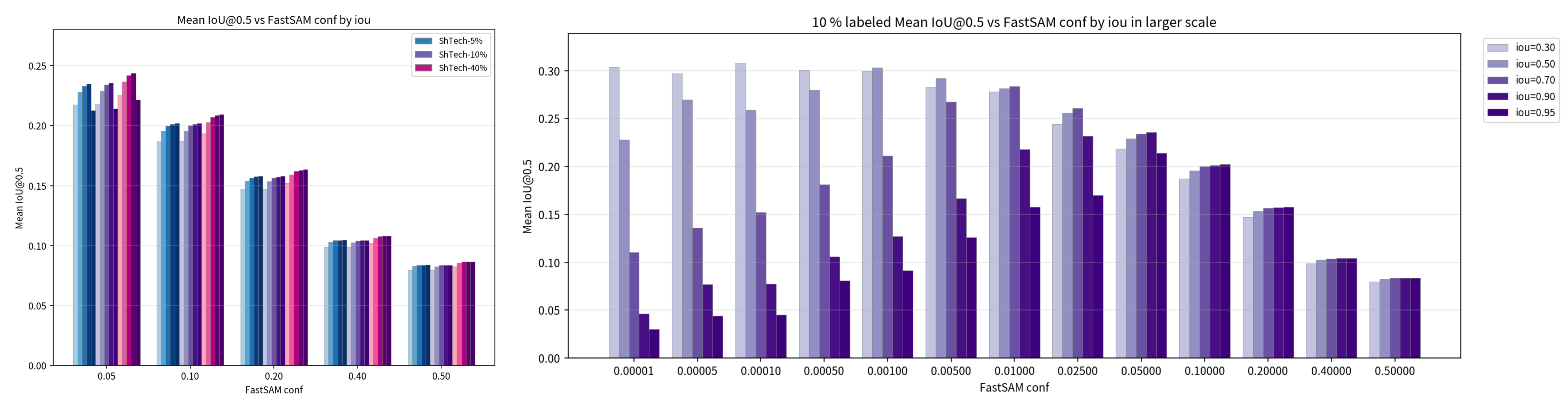}
    \caption{FastSAM Parameters. To obtain competitive $IoU@0.5$ in Point-prompt Instance Segmentation Task, we evaluated $conf$ and $iou$. Left figure illustrates the performance between different protocols are similar, right figure shows 10\% results on a larger scale. Considering both the high quality of the final generated mask and high IoU, we selected $conf=0.5$ and $iou=0.01$ in FastSAM. }
\end{subfigure}
\hfill
\begin{subfigure}{1.0\linewidth}
    \label{fig:XMask inference}
    \centering
    \includegraphics[width=\linewidth]{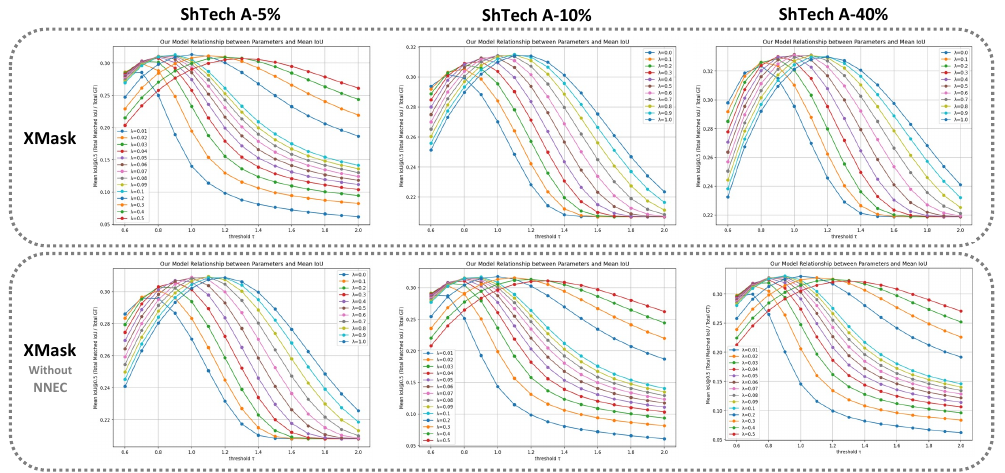}
    \caption{XMask Parameters. The generation of XMask segmentation relate to threshold $\tau_g$ and geometry weight $\lambda$. According to the results, we selected $\tau_g$ = 1.1 and $\lambda$ = 0.8 in NNEC applied method, $\tau_g$ = 1.0 and $\lambda$ = 0.2 in without NNEC to generate Instance Segmentation.}
\end{subfigure}
\end{figure}

Therefore, we experimented with these two parameters to obtain the best segment performance that has the highest IoU@0.5 in our XMask model. As presented in \cref{fig:XMask inference}, three graphs in the first row compare the performance under different settings of XMask, and the result shows that when $\lambda$ = 1.1 and $\tau_g$ = 0.8, our model achieves the best with the highest IoU@0.5. Additionally, when $\lambda$ = 1.0 and $\tau_g$ = 0.2, XMask (without NNEC) shows the highest IoU@0.5. Finally, we chose the settings above to evaluate our model in three datasets.

\textbf{FastSAM.} There are two significant parameters in FastSAM, $conf$ and $iou$, used to segment everything from the input picture. $conf$ (confidence threshold) determines the minimum confidence score required for a predicted mask to be retained Predictions with confidence values lower than the specified $conf$ threshold are discarded. Increasing this value results in fewer but more reliable masks, while decreasing it allows more candidate masks to be kept. $iou$ (Intersection over Union threshold) is used in the NMS process to remove redundant or highly overlapping masks. A higher $iou$ threshold allows more overlapping masks to coexist, whereas a lower threshold resulting in the filtering stricter and more redundant predictions. 

Similarly, we experimented with IoU@0.5 in various FastSAM settings. As shown in \cref{fig:fastsam iou}, we evaluated the performance on the ShTech A dataset. In the left bar graph, we tested the labeled semi-supervised trained model at 5\%, 10\% and 40\%. We then obtained the predicted point location and gave the point prompts to FastSAM. We tested the IoU@0.5 of the FastSAM output without the NNEC constraint against the ground truth. We evaluated conf from 0.05 to 0.5 and iou from 0.3 0.5 0.7 and 0.95, higher values corresponding to darker shades of bar color in the figure. Results show that these three different percentages performance simultaneously in different settings. Subsequently, we just experimented 10\% in a wider range, as in the right figure. When $conf$ is smaller than 0.01 and $iou$ is 0.3 and 0.5, IoU@0.5 changes within a small range. Therefore, considering the high quality and accuracy of the final mask, we selected $conf$ = 0.5 and $iou$ = 0.01 as FastSAM baseline setting.

\section{Additional Experiments of EDP-SAM}

\subsection{hyperparameter in SLIC.} 
\begin{figure}
    \centering
    \includegraphics[width=1.0\linewidth]{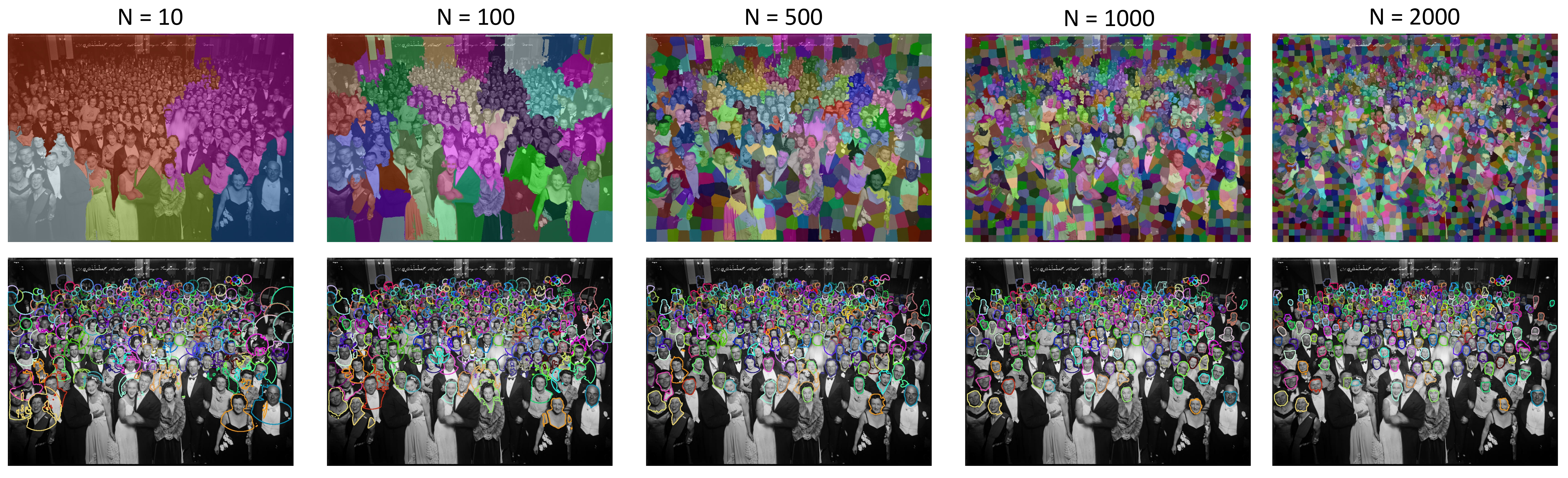}
    \caption{Experiments on number of segment comparison to generate superpixel by SLIC. We implement n = 1000 in our approach, because smaller number leads to inaccurate mask without semantic, and larger number results in broken masks.}
    \label{fig:SLIC}
\end{figure}

Before implementing SLIC to generate superpixel prompt, we tested how superpixel performance changes as \textbf{n segments} various, representing the approximate number in the segmented output.  As shown in \cref{fig:SLIC}, we choose n = 1000 to generate superpixel prompts, finally resulting in accurate and complete masks. Subsequently, we will use a superpixel as one of our prompts given to SAM.

\subsection{Framework of EDP-SAM.} 
\begin{figure}
    \centering
    \includegraphics[width=1.0\linewidth]{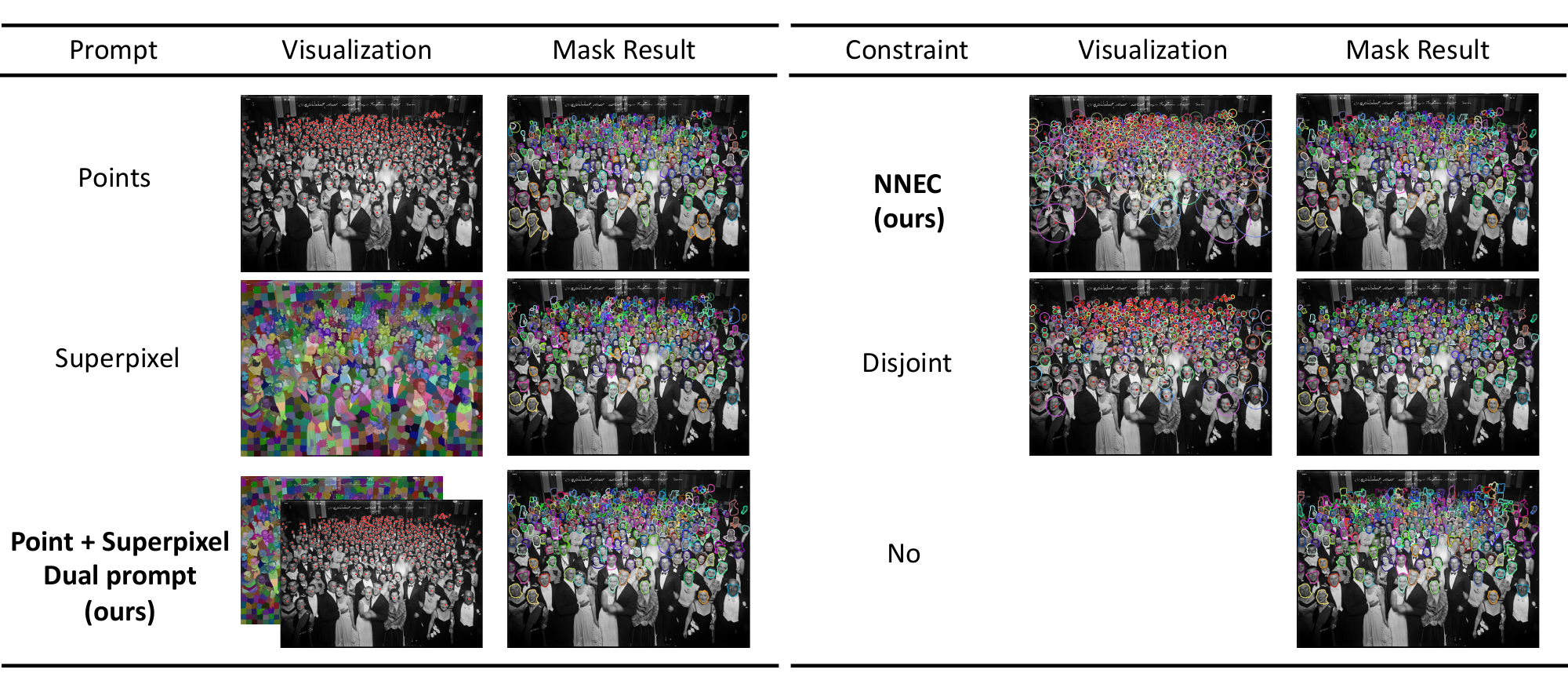}
    \caption{Prompt and Constraint Ablation Study in EDP-SAM. The left table illustrates that our dual-prompt input results in both complete and accurate (in count) masks, and the right table shows the effectiveness of regional constraint and our NNEC circle lead to better masks with complete crowd head.}
    \label{fig:Component in EDP-SAM}
\end{figure}

As mentioned in the main part, our method NNEC point and superpixels dual-prompt SAM (EDP-SAM) consists of two prompts with regional constraint. To illustrate the effictiveness of our design, we show a comparison in \cref{fig:Component in EDP-SAM}. 

In the left table, we compared point prompt, superpixel prompt and our dual-prompt that combines two of them with NNEC constraint. Using only the point prompt leads to broken masks, as shown in the first set of mask results, especially person located near the scene. Conversely, using only the superpixel prompt results in fewer masks than the true count of people in the crowd. However, our proposed dual-prompt approach solves these problems by generating accurate and complete masks using SAM.

In the right table, we analyzed the beneficial effect of NNEC constraint. With point annotations, we initially proposed two methods: exclusion circle with the closest distance and the half of closest distance as radius, which means the latter are disjoint circles. Firstly, as shown in the last row, if there is no regional constraint, mask especially for the identity in the distance will in a bad situation, with strange size. Comparing the first and second rows, disjoint circles perform worse when crowds overlap, but ours will solve this problem. 

\begin{figure}
    \centering
    \includegraphics[width=1.0\linewidth]{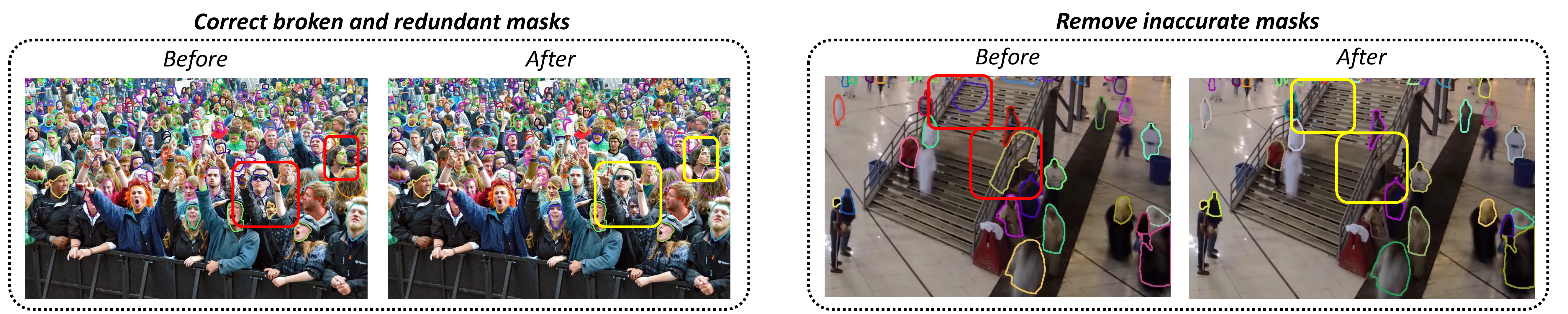}
    \caption{Examples of Artificial Correction in EDP-SAM.}
    \label{fig:mamual correction}
\end{figure}

\textbf{Artificial Correction.} After getting the intersection from SAM output and NNEC constraint, we apply an artificial correction to create a reliable and accurate mask ground truth dataset. As shown in \cref{fig:mamual correction}, there are several problems in the initial intersection results. For example, broken masks with missing items (e.g., sunglasses), incomplete head coverage, or masks without human regions.

% More ablation results are shown in \cref{tab:ablation}.

% \begin{table}[h]
% \centering
% \caption{Additional ablation study.}
% \begin{tabular}{c|c|c}
% Method & MAE & MSE \\
% \hline
% Baseline & 65.3 & 102.1 \\
% Ours & 58.7 & 93.4
% \end{tabular}
% \label{tab:ablation}
% \end{table}

\section{Semi-Supervised Counting Visualization}
\begin{figure}
    \centering
    \includegraphics[width=0.9\linewidth]{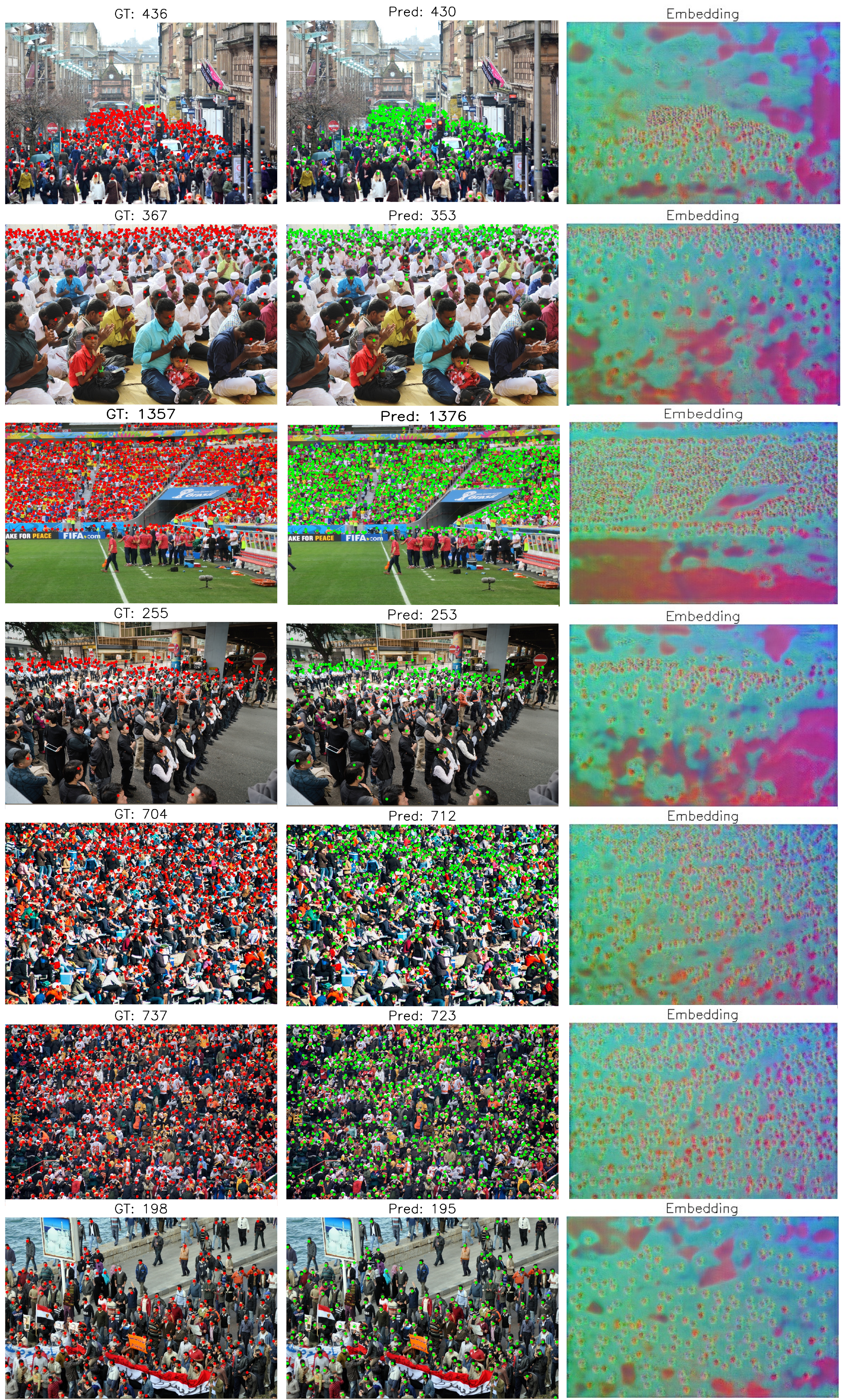}
    \caption{Visualization of Semi-supervised Counting Comparison with GT}
    \label{fig:visualization}
\end{figure}
% \end{document}

\end{document}